\documentclass[10pt,twocolumn,letterpaper]{article}

\usepackage{cvpr}
\usepackage{times,fullpage,times}
\usepackage{url}
\usepackage{epsfig}
\usepackage{graphicx}
\usepackage{amsmath}
\usepackage{amssymb}
\usepackage{caption}
\usepackage{subcaption}
\usepackage{algorithm}
\usepackage{algpseudocode}
\usepackage{pifont}
\usepackage{xcolor}
\usepackage{rotating}
\usepackage{epstopdf}
\usepackage{caption}
\usepackage{subcaption}

\usepackage{amsmath,amsfonts}
\usepackage{amssymb,amsopn}
\usepackage{bm} 

\usepackage{amssymb}
\usepackage{amsmath,amsfonts}
\usepackage{amsthm,amsopn}

\usepackage{bm} 

\newcommand{\vct}[1]{\boldsymbol{#1}} 
\newcommand{\mat}[1]{\boldsymbol{#1}} 
\newcommand{\cst}[1]{\mathsf{#1}}  

\newcommand{\field}[1]{\mathbb{#1}}
\newcommand{\R}{\field{R}} 
\newcommand{\I}{\field{I}} 
\newcommand{\T}{^{\textrm T}} 

\newcommand{\twonorm}[1]{\left\|#1\right\|_2^2}

\newcommand{\ProbOpr}[1]{\mathbb{#1}}

\newcommand{\expect}[2]{%
\ifthenelse{\equal{#2}{}}{\ProbOpr{E}_{#1}}
{\ifthenelse{\equal{#1}{}}{\ProbOpr{E}\left[#2\right]}{\ProbOpr{E}_{#1}\left[#2\right]}}} 
\newcommand{\var}[2]{%
\ifthenelse{\equal{#2}{}}{\ProbOpr{VAR}_{#1}}
{\ifthenelse{\equal{#1}{}}{\ProbOpr{VAR}\left[#2\right]}{\ProbOpr{VAR}_{#1}\left[#2\right]}}} 

\newcommand{\vx}{{\vct{x}}}

\newcommand{\va}{\vct{a}}
\newcommand{\vb}{\vct{b}}

\newcommand{\vv}{\vct{v}}
\newcommand{\vw}{\vct{w}}

\newcommand{\mM}{\mat{M}}

\newcommand{\mI}{\mat{I}}

\newcommand{\mSigma}{\mat{\Sigma}}

\newcommand{\cN}{\cst{N}}

\newcommand{\cD}{\cst{D}}

\newcommand{\cR}{\cst{R}}
\newcommand{\cU}{\cst{U}}
\newcommand{\cS}{\cst{S}}

\newcommand{\eat}[1]{}

\newcommand{\WC}[1]{{\color{blue} WC: #1}}

\usepackage[pagebackref=true,breaklinks=true,colorlinks,bookmarks=false]{hyperref}
\cvprfinalcopy 

\def\cvprPaperID{2047} 

\ifcvprfinal\fi
\begin{document}

\title{Synthesized Classifiers for Zero-Shot Learning}

\author{Soravit Changpinyo$^*$, Wei-Lun Chao\thanks{\hspace{4pt}Equal contributions}\\
U. of Southern California\\
Los Angeles, CA\\
{\tt\small schangpi, weilunc@usc.edu}
\and
Boqing Gong\\
U. of Central Florida\\
Orlando, FL\\
{\tt\small bgong@crcv.ucf.edu}
\and
Fei Sha\\
U. of California\\
Los Angeles, CA\\
{\tt\small feisha@cs.ucla.edu}
}

%

\newcommand{\fix}{\marginpar{FIX}}
\newcommand{\new}{\marginpar{NEW}}

\maketitle

\begin{abstract}
Given semantic descriptions of object classes, zero-shot learning aims to accurately recognize objects of the \emph{unseen} classes, from which no examples are available at the training stage, by associating them to the \emph{seen} classes, from which labeled examples are provided.
We propose to tackle this problem from the perspective of manifold learning. 
Our main idea is to align the semantic space that is derived from external information to the model space that concerns itself with recognizing visual features.
To this end, we introduce a set of ``phantom'' object classes whose coordinates live in both the semantic space and the model space.
Serving as bases in a dictionary, they can be optimized from labeled data such that the \emph{synthesized} real object classifiers achieve optimal discriminative performance. We demonstrate superior accuracy of our approach over the state of the art on four benchmark datasets for zero-shot learning, including the full ImageNet Fall 2011 dataset with more than 20,000 unseen classes.
\end{abstract}


\section{Introduction}

Visual recognition has made significant progress due to the widespread use of  deep learning architectures~\cite{KrizhevskySH12, SzegedyLJSRAEVR14} that are optimized on large-scale datasets of human-labeled images~\cite{ILSVRC15}.  Despite the exciting advances, to recognize objects ``in the wild'' remains a daunting challenge. Many objects follow  a long-tailed distribution: in contrast to common objects such as household items, they do not occur frequently enough for us to collect and label a large set of representative exemplar images. 

For example, this challenge is especially crippling for fine-grained object recognition (classifying species of birds,  designer products, etc.). Suppose we want to carry a visual search of ``Chanel Tweed Fantasy Flap Handbag''.  While handbag, flap, tweed, and Chanel are popular accessory, style, fabric, and brand, respectively, the combination of them is rare --- the query generates about 55,000 results on Google search with a small number of images. The amount of labeled images is thus far from enough for directly building a high-quality classifier, unless we treat this category as a composition of attributes, for each of which more training data can be easily acquired~\cite{LampertNH14}.

It is thus imperative to develop methods for zero-shot learning, namely, to expand  classifiers and the space of possible labels beyond \emph{seen} objects, of which we have access to the labeled images for training, to \emph{unseen} ones, of which no labeled images are available~\cite{LampertNH14, PalatucciPHM09}.  To this end, we need to address two key interwoven challenges~\cite{PalatucciPHM09}: (1) how to relate unseen classes to seen ones and (2) how to attain optimal discriminative performance on the unseen classes even though we do not have their labeled data.

To address the first challenge, researchers have been using visual attributes~\cite{FarhadiEHF09, LampertNH09, ParikhG11} and word vectors~\cite{FromeCSBDRM13, MikolovCCD13, SocherGMN13} to  associate  seen and unseen classes. We call them the semantic embeddings of objects. Much work takes advantage of such embeddings directly as middle layers between input images and output class labels~\cite{AkataPHS13,FromeCSBDRM13,JayaramanG14,LampertNH14,Li_2015_ICCV,li2014attributes,NorouziMBSSFCD14,Bernardino15,SocherGMN13}, whereas others derive new representations from the embeddings using, for example, Canonical Correlation Analysis (CCA)~\cite{FuHXFG14,fu2015pami,lu2015unsupervised}\eat{linear (or bilinear) transformation~\cite{AkataPHS13,Bernardino15,Li_2015_ICCV}} or sparse coding~\cite{Kodirov_2015_ICCV, zhang2015classifying, zhang2015zero}. For the second challenge, the hand-designed probabilistic models in~\cite{LampertNH14} have been competitive baselines. More recent studies show that nearest neighbor classifiers in the semantic space are very effective~\cite{FromeCSBDRM13,FuHXFG14,fu2015pami,fu2015zero,NorouziMBSSFCD14}.\eat{This is further strengthened by distance metric learning~\cite{fu2015zero}.} Finally, classifiers for the unseen classes can directly be constructed in the input feature space~\cite{AkataPHS13,ElhoseinySE13,Ba_2015_ICCV,MensinkGS14,Yang2015unified,zhang2015zero}.

\begin{figure*}
\centering
\includegraphics[width=0.9\textwidth]{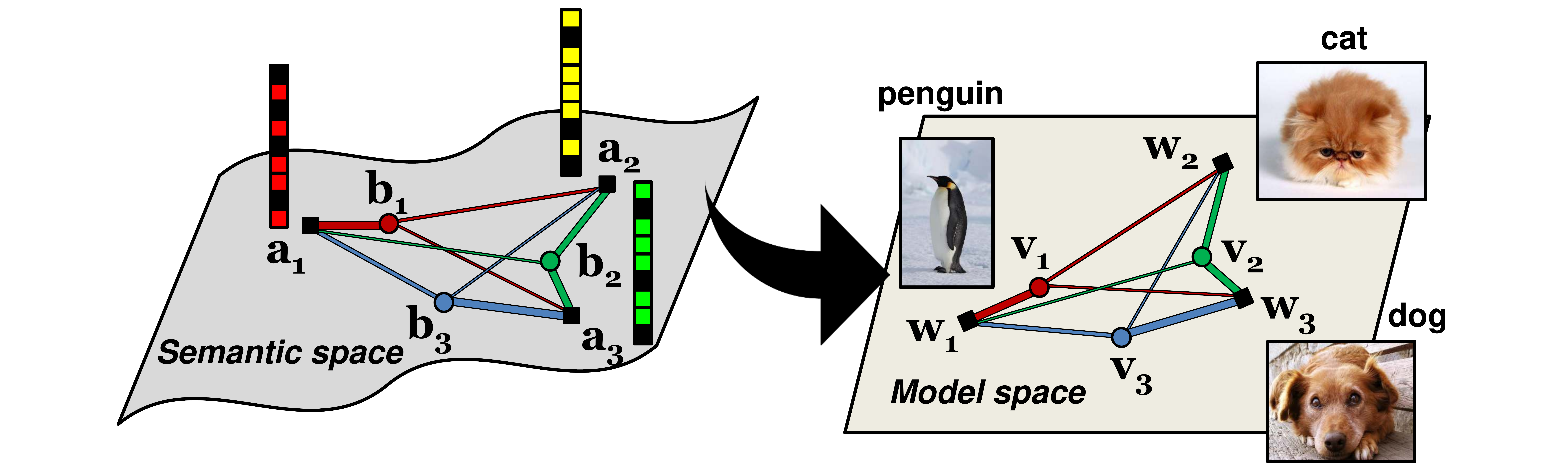}
\vspace{-10pt}
\caption{\small Illustration of our method for zero-shot learning. Object classes live in two spaces. They are characterized in the semantic space with semantic embeddings ($\va$s) such as attributes and word vectors of their names. They are also represented as models for visual recognition ($\vw$s) in the model space. In both spaces, those classes form weighted graphs. The main idea behind our approach is that these two spaces should be aligned. In particular, the coordinates in the model space should be the projection of the graph vertices from the semantic space to the model space --- preserving class relatedness encoded in the graph. We introduce adaptable phantom classes ($\vb$ and $\vv$) to connect seen and unseen classes --- classifiers for the phantom classes are bases for synthesizing classifiers for real classes. In particular, the synthesis takes the form of convex combination. }
\label{fConcept}
\vskip -1em
\end{figure*}

In this paper, we tackle these two challenges with ideas from manifold learning~\cite{belkin2003laplacian, hinton2002stochastic}, converging to a two-pronged approach. We view object classes in a semantic space as a  weighted graph where the nodes correspond to object class names and the weights of the edges represent how they are related. Various information sources can be used to infer the weights --- human-defined attributes or word vectors learnt from language corpora. On the other end, we view models for recognizing visual images of those classes as if they live in a space of models.  In particular, the parameters for each object model are nothing but coordinates in this model space whose geometric configuration also reflects the relatedness among objects. Fig.~\ref{fConcept} illustrates this idea conceptually.

But \emph{how do we align the semantic space and the model space}? The semantic space coordinates of objects are designated or derived based on external information (such as textual data) that do not directly examine visual appearances at the lowest level, while the model space concerns itself largely for recognizing low-level visual features. To align them, we view the coordinates in the model space as the projection of the vertices on the graph from the semantic space --- there is a wealth of literature on manifold learning for computing (low-dimensional) Euclidean space embeddings from the weighted graph, for example, the well-known algorithm of Laplacian eigenmaps~\cite{belkin2003laplacian}.

To adapt the embeddings (or the coordinates in the model space) to data, we introduce a set of \emph{phantom} object classes --- the coordinates of these classes in both the semantic space and the model space are adjustable and optimized such that the resulting model for the real object classes achieve the best performance in discriminative tasks. However, as their names imply, those phantom classes do not correspond to and are not optimized to recognize any real classes directly. For mathematical convenience, we parameterize the weighted graph in the semantic space with the phantom classes in such a way that the model for any real class is a convex combinations of the coordinates of those phantom classes. In other words, the ``models'' for the phantom classes can also be interpreted as  bases (classifiers) in a dictionary from which a large number of classifiers for real classes can be synthesized via convex combinations. In particular, when we need to construct a classifier for an unseen class, we will compute the convex combination coefficients from this class's semantic space coordinates and use them to form the corresponding classifier.

To summarize, our main contribution is a novel idea to cast the challenging problem of recognizing unseen classes as learning manifold embeddings from graphs composed of object classes.  As a concrete realization of this idea, we show how to parameterize the graph with the locations of the phantom classes, and how to derive embeddings (i.e., recognition models) as convex combinations of base classifiers.   Our empirical studies extensively test our synthesized classifiers  on four benchmark datasets for zero-shot learning, including the full  ImageNet Fall 2011 release~\cite{deng2009imagenet} with 20,345 unseen classes. The  experimental results are very encouraging; the synthesized classifiers outperform several state-of-the-art methods, including attaining better or matching performance of Google's ConSE algorithm~\cite{NorouziMBSSFCD14} in the large-scale setting.

The rest of the paper is organized as follows. We give an overview of relevant literature in Section~\ref{sRelated}, describe our approach in detail in Section~\ref{sApproach}, demonstrate its effectiveness in Section~\ref{sExp}, and conclude in Section~\ref{sDiscuss}.

\section{Related Work}
\label{sRelated}

In order to transfer knowledge between classes, zero-shot learning relies on semantic embeddings of class labels, including attributes (both manually defined ~\cite{AkataPHS13, LampertNH14, WangJ13} and discriminatively learned~\cite{AlHalahS15, YuCFSC13}), word vectors~\cite{FromeCSBDRM13, NorouziMBSSFCD14, RohrbachSSGS10, SocherGMN13}, knowledge mined from the Web~\cite{ElhoseinySE13, MensinkGS14, RohrbachSS11, RohrbachSSGS10}, or a combination of several embeddings~\cite{AkataRWLS15, FuHXFG14, fu2015zero}.

Given semantic embeddings, existing approaches to zero-shot learning mostly fall into embedding-based and similarity-based methods. In the embedding-based approaches, one first maps the input image representations to the semantic space, and then determines the class labels in this space by various relatedness measures implied by the class embeddings~\cite{AkataPHS13, AkataRWLS15, FromeCSBDRM13, FuHXFG14, fu2015zero, Kodirov_2015_ICCV, LampertNH14, li2014attributes, NorouziMBSSFCD14, SocherGMN13, WangJ13}. Our work as well as some recent work combine these two stages~\cite{AkataPHS13, AkataRWLS15, FromeCSBDRM13, Bernardino15, WangJ13, zhang2015classifying, zhang2015zero}, leading to a unified framework empirically shown to have, in general, more accurate predictions. In addition to directly using fixed semantic embeddings, some work maps them into a different space through CCA~\cite{FuHXFG14,fu2015pami,lu2015unsupervised}\eat{linear (or bilinear)transformation~\cite{AkataPHS13,Li_2015_ICCV,Bernardino15}} and sparse coding~\cite{Kodirov_2015_ICCV, zhang2015classifying, zhang2015zero}. 

In the similarity-based approaches, in contrast, one builds the classifiers for unseen classes by relating them to seen ones via class-wise similarities~\cite{ElhoseinySE13, fu2015zero,Gavves_2015_ICCV, MensinkGS14,RohrbachSS11,RohrbachSSGS10}. Our approach shares a similar spirit to these models but offers richer modeling flexibilities thanks to the introduction of phantom classes.

Finally, our convex combination of base classifiers for synthesizing real classifiers can also be motivated from multi-task learning with shared representations~\cite{ArgyriouEP08}. While labeled examples of each task are required in~\cite{ArgyriouEP08}, our method has no access to \emph{data} of the unseen classes.

\section{Approach}
\label{sApproach}

We describe our methods for addressing zero-shot learning where the task is to classify images from unseen classes into the label space of unseen classes.  

\paragraph{Notations} Suppose we have training data $\mathcal{D}= \{(\vx_n\in \R^{\cD},y_n)\}_{n=1}^\cN$ with the labels coming from the label space of \emph{seen} classes $\mathcal{S} = \{1,2,\cdots,\cS\}$.  Denote by $\mathcal{U} = \{\cS+1,\cdots,\cS+\cU\}$ the label space of \emph{unseen} classes.

We focus on linear classifiers in the visual feature space $\R^\cD$ that assign a label $\hat{y}$ to a data point $\vx$ by
\begin{align}
\hat{y} = \arg\max_{c} \quad \vw_c\T\vx, 
\label{ePredict}
\end{align}
where $\vw_c \in \R^{\cD}$, although our approach can be readily extended to nonlinear settings by the kernel trick~\cite{scholkopf2002learning}.

\subsection{Main idea} \label{sConventional}
 
\paragraph{Manifold learning } The main idea behind our approach is shown by the conceptual diagram in Fig.~\ref{fConcept}. Each class $c$ has a coordinate $\va_c$ and they live on a manifold in the semantic embedding space. In this paper, we explore two types of such spaces:  attributes~\cite{LampertNH14, WahCUB_200_2011} and class name embeddings via word vectors~\cite{MikolovSCCD13}.  We use attributes in this text to illustrate the idea and in the experiments we test our approach on both types. 

Additionally, we introduce a set of \emph{phantom} classes associated with semantic embeddings $\vb_r, r =1, 2, \ldots, \cR$.  We stress that they are  phantom as they themselves do \textbf{not} correspond to any real objects --- they are introduced to increase the modeling flexibility, as shown below.

The real and phantom classes form a weighted bipartite graph, with the weights defined as
\begin{align}
s_{cr} = \frac{\exp\{-d(\va_c,\vb_r)\}}{\sum_{r=1}^{\cR}\exp\{-d(\va_c,\vb_r)\}}
\label{simform}
\end{align}
to correlate a real class $c$ and a phantom class $r$, where 
\begin{align}
d(\va_c,\vb_r)=(\va_c-\vb_r)^T\mSigma^{-1}(\va_c-\vb_r), \label{eDist}
\end{align}
and $\mSigma^{-1}$ is a parameter that can be learned from data, modeling the correlation among attributes.   For simplicity, we set $\mSigma = \sigma^2\mat{I}$ and tune the scalar free hyper-parameter  $\sigma$ by cross-validation. The more general Mahalanobis metric can be used and we propose one way of learning such metric as well as demonstrate its effectiveness in the Suppl. 

The specific form of defining the weights is motivated by several manifold learning methods such as SNE~\cite{hinton2002stochastic}. In particular, $s_{cr}$ can be interpreted as the conditional probability of observing class $r$ in the neighborhood of class $c$.  However, other forms can be explored and are left for future work.

In the model space, each real class is associated with a classifier $\vw_c$ and the phantom class $r$ is associated with a virtual classifier $\vv_r$. We align the semantic and the model spaces by viewing $\vw_c$ (or $\vv_r$) as the embedding of the weighted graph. In particular, we appeal to the idea behind Laplacian eigenmaps~\cite{belkin2003laplacian}, which seeks the embedding that maintains the graph structure as much as possible; equally, the distortion error
\[
\min_{\vw_c, \vv_r} \| \vw_c - \sum_{r=1}^{\cR} s_{cr} \vv_r\|_2^2
\]
is minimized.  This objective has an analytical solution 
\begin{align}
\vw_c = \sum_{r=1}^\cR s_{cr}\vv_r, \quad \forall\, c\in\mathcal{T}=\{1,2,\cdots,\cS+\cU\} \label{eCombine}
\end{align}
In other words, the solution gives rise to the idea of \emph{synthesizing classifiers} from those virtual classifiers $\vv_r$. For conceptual clarity, from now on we refer to $\vv_r$ as base classifiers in a dictionary from which new classifiers can be synthesized. We identify several advantages. First,  we could construct an infinite number of classifiers as long as we know how to compute $s_{cr}$. Second, by  making $\cR\ll\cS$, the formulation can significantly reduce the learning cost as we only need to learn $\cR$ base classifiers.

\subsection{Learning phantom classes}
\label{sLearnPhantom}
\paragraph{Learning base classifiers} We learn the  base classifiers $\{\vv_r\}_{r=1}^\cR$ from the training data (of the seen classes only). We experiment with two settings. To learn one-versus-other classifiers, we optimize,
\begin{align}
&\min_{\vv_1,\cdots,\vv_\cR} \sum_{c=1}^{\cS}\sum_{n=1}^\cN \ell({\vx_n}, \I_{y_n,c}; {\vw_c}) + \frac{\lambda}{2} \sum_{c=1}^{\cS} \twonorm{\vw_c}, \label{eObj} \\
&\mathsf{s.t.}\quad \vw_c = \sum_{r=1}^\cR s_{cr}\vv_r, \quad \forall\, c\in\mathcal{T}=\{1,\cdots,\cS\}  \notag
\end{align}
where $\ell(\vx, y; \vw)=\max(0,1-y\vw\T\vx)^2$ is the squared hinge loss. The indicator $\I_{y_n,c}\in\{-1,1\}$ denotes whether or not $y_n=c$. 
Alternatively, we apply the Crammer-Singer multi-class SVM loss~\cite{CrammerS02}, given by

\begin{align*}
&\ell_\text{cs}(\vx_n,\hspace{2pt}y_n; \{\vw_c\}_{c=1}^{\cS}) \notag \\
= &\max (0, \max_{c\in \mathcal{S}-\{y_n\}} \Delta(c,y_n) + {\vw_c}\T{\vx_n} - {\vw_{y_n}}\T{\vx_n}),
\end{align*}
We have the standard Crammer-Singer loss when the structured loss $\Delta(c,y_n)=1$ if $c\neq y_n$, which, however, ignores the semantic relatedness between classes. We additionally  use the $\ell_2$ distance for the structured loss $\Delta(c,y_n)=\twonorm{\va_c-\va_{y_n}}$ to exploit  the class relatedness in our experiments.
These two learning settings have separate strengths and weaknesses in empirical studies.

\paragraph{Learning semantic embeddings}\label{sEmbeddingPhantom}  The weighted graph eq.~(\ref{simform})  is also parameterized by adaptable embeddings of the phantom classes $\vb_r$. For this work, however, for simplicity, we assume that each of them is a sparse linear combination of the seen classes' attribute vectors:
\begin{equation}
\vb_r = \sum_{c=1}^\cS\beta_{rc}\va_c,\forall r\in\{1,\cdots,\cR\}, \notag
\end{equation}
Thus, to optimize those embeddings, we solve the following optimization problem
\begin{align*}
&\min_{\{\vv_r\}_{r=1}^\cR,
\{\beta_{rc}\}_{r,c=1}^{\cR, \cS}} \sum_{c=1}^{\cS}\sum_{n=1}^\cN \ell({\vx_n}, \I_{y_n,c}; {\vw_c})\\
&
+\frac{\lambda}{2} \sum_{c=1}^{\cS} \twonorm{\vw_c}\label{eObj_att}
+\eta\sum_{r,c=1}^{\cR,\cS}|\beta_{rc}|
+\frac{\gamma}{2}\sum_{r=1}^{\cR} (\twonorm{\vb_r}-h^2)^2\notag,\\
&\mathsf{s.t.}\quad \vw_c = \sum_{r=1}^\cR s_{cr}\vv_r, \quad \forall\, c\in\mathcal{T}=\{1,\cdots,\cS\}  \notag,
\end{align*}
where $h$ is a predefined scalar equal to the norm of real attribute vectors (i.e., 1 in our experiments since we perform $\ell_2$ normalization).
Note that in addition to learning $\{\vv_r\}_{r=1}^\cR$, we learn combination weights $\{\beta_{rc}\}_{r, c=1}^{\cR, \cS}.$ 
Clearly, the constraint together with the third term in the objective encourages the sparse linear combination of the seen classes' attribute vectors. The last term in the objective demands that the norm of $\vb_r$ is not too far from the norm of $\va_c$.

We perform alternating optimization for minimizing the objective function with respect to $\{\vv_r\}_{r=1}^\cR$ and $\{\beta_{rc}\}_{r,c=1}^{\cR, \cS}$. While this process is nonconvex, there are useful heuristics to initialize the optimization routine.  For example, if $\cR = \cS$, then the simplest setting is to let $\vb_r = \va_r$ for $r = 1, \ldots, \cR$. If  $\cR\le\cS$, we can let them be (randomly) selected from the seen classes' attribute vectors $\{\vb_1, \vb_2, \cdots, \vb_\cR\} \subseteq \{\va_1, \va_2, \cdots, \va_\cS\}$, or first perform clustering on $\{\va_1, \va_2, \cdots, \va_\cS\}$ and then let each $\vb_r$ be a combination of the seen classes' attribute vectors in cluster $r$. If $\cR > \cS$, we could use a combination of the above two strategies.  We describe in more detail how to optimize and cross-validate hyperparameters in the Suppl.

\subsection{Comparison to several existing methods} We contrast our approach to some existing methods. \cite{MensinkGS14} combines \textbf{pre-trained} classifiers of seen classes to construct new classifiers. To estimate the semantic embedding (e.g., word vector) of a test image, \cite{NorouziMBSSFCD14} uses the decision values of pre-trained classifiers of seen objects to weighted average the corresponding semantic embeddings. Neither of them has the notion of base classifiers, which we introduce for constructing the classifiers and nothing else. We thus expect them to be more effective in transferring knowledge between seen and unseen classes than overloading the pretrained and fixed classifiers of the seen classes for dual duties. We note that~\cite{AkataPHS13} can be considered as a special case of our method. In~\cite{AkataPHS13}, each attribute corresponds to a base and each ``real'' classifier corresponding to an actual object is represented as a linear combination of those bases, where the weights are the real objects' ``descriptions'' in the form of attributes.  This modeling is limiting as the number of bases is fundamentally limited by the number of attributes. Moreover, the model is strictly a subset of our model.\footnote{For interested readers, if we set the number of attributes as the number of phantom classes (each $\vb_r$ is the one-hot representation of an attribute), and use  Gaussian kernel with anisotropically diagonal covariance matrix in eq.~(\ref{eDist}) with properly set bandwidths (either very small or very large) for each attribute, we will recover the formulation in \cite{AkataPHS13} when the bandwidths tend to zero or infinity.} Recently, \cite{zhang2015classifying, zhang2015zero} propose similar ideas of aligning the visual and semantic spaces but take different approaches from ours.

\section{Experiments} \label{sExp}
We evaluate our methods and compare to existing state-of-the-art models on several benchmark datasets. While there is a large degree of variations in the current empirical studies  in terms of datasets, evaluation protocols, experimental settings, and implementation details, we strive to provide a comprehensive comparison to as many methods as possible, not only citing the published results but also reimplementing some of those methods to exploit several crucial insights we have discovered in studying our methods.

We summarize our main results in this section. More extensive details are reported in the Suppl. We provide not only comparison in recognition accuracy but also analysis in an effort to understand the sources of better performance.

\subsection{Setup}
\paragraph{Datasets} 
We use four benchmark datasets in our experiments: the \textbf{Animals with Attributes (AwA)}~\cite{LampertNH14},  \textbf{CUB-200-2011 Birds (CUB)}~\cite{WahCUB_200_2011}, \textbf{SUN Attribute (SUN)}~\cite{PattersonH14}, and the \textbf{ImageNet} (with full 21,841 classes)~\cite{ILSVRC15}. Table~\ref{tDatasets} summarizes their key characteristics. The Suppl. provides more details.

\begin{table}
\centering
{\small
\caption{\small Key characteristics of studied datasets}
\label{tDatasets}
\begin{tabular}{c|c|c|c}
Dataset &  \# of seen & \# of unseen  & Total \# \\ 
name & classes & classes & of images\\ \hline
AwA$^\dagger$ & 40  & 10 & 30,475\\ \hline
CUB$^\ddagger$ & 150 & 50 & 11,788\\ \hline
SUN$^\ddagger$ & 645/646  & 72/71 &  14,340\\ \hline
ImageNet$^\S$ & 1,000 & 20,842 & 14,197,122\\ \hline
\end{tabular}
\begin{flushleft}
\vskip -.5em
$^\dagger$: Following the prescribed split in ~\cite{LampertNH14}.\\
$^\ddagger$: 4 (or 10, respectively) random splits, reporting average.\\
$^\S$: Seen and unseen classes from ImageNet ILSVRC 2012 1K~\cite{ILSVRC15} and Fall 2011 release~\cite{deng2009imagenet,FromeCSBDRM13, NorouziMBSSFCD14}.
\end{flushleft}
}
\vskip -1em
\end{table}

\paragraph{Semantic spaces} For the classes in \textbf{AwA}, we use 85-dimensional binary or continuous attributes~\cite{LampertNH14}, as well as the 100 and 1,000 dimensional word vectors~\cite{MikolovCCD13}, derived from their class names and  extracted by Fu et al.~\cite{FuHXFG14, fu2015pami}. For \textbf{CUB} and \textbf{SUN}, we use 312 and 102 dimensional continuous-valued attributes, respectively. We also thresh them at the global means to obtain binary-valued attributes, as suggested in~\cite{LampertNH14}. Neither datasets have word vectors for their class names. For \textbf{ImageNet}, we train a skip-gram language model~\cite{MikolovCCD13, MikolovSCCD13} on the latest Wikipedia dump corpus\footnote{\url{http://dumps.wikimedia.org/enwiki/latest/enwiki-latest-pages-articles.xml.bz2} on September 1, 2015} (with more than 3 billion words) to extract a 500-dimensional word vector for each class.  Details of this training are in the Suppl. We ignore classes without word vectors in the experiments, resulting in 20,345 (out of 20,842) unseen classes.   For both the continuous attribute vectors and the word vector embeddings of the class names, we normalize them to have unit $\ell_2$ norms unless stated otherwise. 

\paragraph{Visual features} Due to variations in features being used in literature, it is impractical  to try all possible combinations of features and methods. Thus, we make a major distinction in  using shallow features (such as color histograms, SIFT, PHOG, Fisher vectors)~\cite{AkataPHS13,AkataRWLS15,Jayaraman14,LampertNH14,RohrbachSSGS10,WangJ13} and deep learning features in several recent studies of zero-shot learning. Whenever possible, we use (shallow) features provided by those datasets or prior studies. For comparative studies, we also extract the following deep features: AlexNet~\cite{KrizhevskySH12} for \textbf{AwA} and \textbf{CUB} and GoogLeNet~\cite{SzegedyLJSRAEVR14} for all datasets (all extracted with  the Caffe package~\cite{jia2014caffe}). For AlexNet, we use the 4,096-dimensional activations of the penultimate layer (fc7) as features. For GoogLeNet, we take the 1,024-dimensional activations of the pooling units, as in~\cite{AkataRWLS15}.  Details on how to extract those features are in the Suppl.

\paragraph{Evaluation protocols} For \textbf{AwA}, \textbf{CUB}, and \textbf{SUN}, we use the (normalized, by class-size) multi-way classification accuracy, as in previous work. Note that the accuracy is always computed on images from unseen classes.  

Evaluating zero-shot learning on the large-scale \textbf{ImageNet} requires substantially different components from evaluating on the other three datasets. First, two evaluation metrics are used, as in \cite{FromeCSBDRM13}: Flat hit@K (F@K) and Hierarchical precision@K (HP@K).  

F@K is defined as the percentage of test images for which the model returns the  true label in its top K predictions. Note that, F@1 is the multi-way classification accuracy. HP@K takes into account the hierarchical organization of object categories. For each true label, we generate a ground-truth list of K closest categories in the hierarchy and compute the degree of overlapping (i.e., precision) between the ground-truth and the model's top K predictions.  For the detailed description of this metric, please see the Appendix of \cite{FromeCSBDRM13} and the Suppl.

Secondly, following the procedure in \cite{FromeCSBDRM13, NorouziMBSSFCD14}, we evaluate  on three scenarios of increasing difficulty:
\begin{itemize}
\item \emph{2-hop} contains 1,509 unseen classes that are within two tree hops of the seen 1K classes according to the ImageNet label hierarchy\footnote{\url{http://www.image-net.org/api/xml/structure_released.xml}}.
\item \emph{3-hop} contains 7,678 unseen classes that are within three tree hops of seen classes.
\item \emph{All} contains all 20,345 unseen classes in the ImageNet 2011 21K dataset that are not in the ILSVRC 2012 1K dataset.
\end{itemize}
The numbers of unseen classes are slightly different from what are used in~\cite{FromeCSBDRM13, NorouziMBSSFCD14} due to the missing semantic embeddings (i.e., word vectors) for certain class names.  

In addition to reporting published results, we have also reimplemented the state-of-the-art method ConSE~\cite{NorouziMBSSFCD14} on this dataset, introducing a few improvements. Details are in the Suppl.

\paragraph{Implementation details} We cross-validate all hyperparameters --- details are in the Suppl.
For convenience, we set the number of phantom classes $\cR$ to be the same as the number of seen classes $\cS$, and set $\vb_r = \va_c$ for $r=c$. We also experiment setting different $\cR$ and learning $\vb_r$.
Our study (cf. Fig.~\ref{fNumPhantom}) shows that when $\cR$ is about 60\% of $\cS$, the performance saturates. We denote the three variants of our methods in constructing classifiers (Section~\ref{sLearnPhantom}) by 
Ours$^{\textrm{o-vs-o}}$ (one-versus-other), Ours$^{\textrm{cs}}$ (Crammer-Singer)  and Ours$^{\textrm{struct}}$ (Crammer-Singer with structured loss).

\begin{table}
\centering
\small {
\caption{\small Comparison between our results and the previously \emph{published} results in multi-way classification accuracies (in \%) on the task of zero-shot learning. For each dataset, the best is in red and the 2nd best is in blue.} \label{tbMain}
\vskip 0.25em
\small
\begin{tabular}{c|c|c|c|c}
\text{Methods}& \textbf{AwA} & \textbf{CUB} & \textbf{SUN} &\textbf{ImageNet}\\ \hline
\text{DAP} \cite{LampertNH14}& \textbf{\emph{41.4} } &  - & \textbf{\emph{22.2}} &-  \\
\text{IAP} \cite{LampertNH14}&   \textbf{\emph{42.2} } &  -  &  \textbf{\emph{18.0}}&-  \\
\text{BN} \cite{WangJ13}  & \textbf{\emph{43.4}}   & -   & -  &- \\
\text{ALE} \cite{AkataPHS13}& \textbf{\emph{37.4}} &\textbf{\emph{18.0}}$^\dagger$& - &- \\
\text{SJE} \cite{AkataRWLS15}  &   \textbf{\emph{66.7}} & \textbf{\emph{50.1}}$^\dagger$ & -&- \\
\text{ESZSL} \cite{Bernardino15} &  \textbf{\emph{49.3}} &- & - &-\\ 
\text{ConSE}\cite{NorouziMBSSFCD14} & - & - & - & {\color{blue}\textbf{\emph{1.4}}}\\ 
\text{SSE-ReLU}~\cite{zhang2015zero}$^\star$ & {\color{blue}\textbf{\emph{76.3}}} & \textbf{\emph{30.4}}$^\dagger$ & - & - \\
\cite{zhang2015classifying}$^\star$ & {\color{red}\textbf{\emph{80.5}}} & \textbf{\emph{42.1}}$^\dagger$ & - & - \\ \hline
{Ours$^\textrm{o-vs-o}$} &  69.7 & {\color{blue}53.4} & {\color{red}62.8}& {\color{blue}1.4}\\
{Ours$^\textrm{cs}$}\hspace{10.5pt} & 68.4 & 51.6 & 52.9 & -\\
{Ours$^\textrm{struct}$}\hspace{2pt} & 72.9 & {\color{red}54.7} & {\color{blue}62.7} & {\color{red}1.5}\\ \hline
\end{tabular}
\begin{flushleft}
$^\dagger$: Results reported\eat{by the authors} on a particular seen-unseen split\eat{that is not publicly available}. \\
$^\star$: Results were just brought to our attention. Note that VGG \cite{Simonyan15c} instead of GoogLeNet features were used, improving on \textbf{AwA} but worsening on \textbf{CUB}. Our results using VGG will appear in a longer version of this paper. \\
\end{flushleft}
}
\vskip -1em
\end{table}  

\subsection{Experimental results} \label{4_2}

\subsubsection{Main results} 

Table~\ref{tbMain} compares the proposed methods to the state-of-the-art from the previously published results on benchmark datasets.
While there is a large degree of variations in implementation details, the main observation is that our methods attain the best performance in most scenarios. In what follows, we analyze those results in detail. 

We also point out that the settings in some existing work are highly different from ours; we do not include their results in the main text for fair comparison \cite{AlHalahS15,FuHXFG14,fu2015pami,fu2015zero,JayaramanG14,Kodirov_2015_ICCV,Li_2015_ICCV,YuCFSC13} --- but we include them  in the Suppl. In some cases, even with additional data and attributes, those methods underperform ours.

\begin{table*}
\centering
\small 
\caption{\small Comparison between results by ConSE and our method on \textbf{ImageNet}. For both types of metrics, the higher the better.} 
\label{tbImagenet}
\small
\begin{tabular}{c|c|ccccc|cccc}
\text{Scenarios} & \text{Methods} & \multicolumn{5}{|c|}{Flat Hit@K} & \multicolumn{4}{|c}{Hierarchical precision@K}\\ \cline{3-11}
& K= & \text{1} & \text{2} & \text{5} & \text{10} & \text{20} & \text{2} & \text{5} & \text{10} & \text{20} \\ \hline 
\emph{2-hop} 
						 & ConSE~\cite{NorouziMBSSFCD14}\hspace{4pt} & 9.4 & 15.1 & 24.7 & 32.7 & 41.8 & 21.4 & 24.7 & 26.9 & 28.4 \\
						 & ConSE by us & 8.3 & 12.9 & 21.8 & 30.9 & 41.7 & 21.5 & 23.8 & 27.5 & 31.3 \\ \cline{2-11}
						 & Ours$^\textrm{o-vs-o}$ & {\color{red}10.5} & {\color{red}16.7} & {\color{red}28.6} & {\color{red}40.1} & {\color{red}52.0} & {\color{red}25.1} & {\color{red}27.7} & {\color{red}30.3} & {\color{red}32.1} \\
						 & Ours$^\textrm{struct}$\hspace{2pt} & 9.8 & 15.3 & 25.8 & 35.8 & 46.5 & 23.8 & 25.8 & 28.2 & 29.6 \\
\hline
\emph{3-hop} 
						 & ConSE~\cite{NorouziMBSSFCD14}\hspace{4pt} & 2.7 & 4.4 & 7.8 & 11.5 & 16.1 & 5.3 & 20.2 & 22.4 & 24.7 \\
						 & ConSE by us & 2.6 & 4.1 & 7.3 & 11.1 & 16.4 & 6.7 & 21.4 & 23.8 & 26.3 \\ \cline{2-11}
						 & Ours$^\textrm{o-vs-o}$ & {\color{red}2.9} & {\color{red}4.9} & {\color{red}9.2} & {\color{red}14.2} & {\color{red}20.9} & 7.4 & {\color{red}23.7} & {\color{red}26.4} & {\color{red}28.6} \\
						 & Ours$^\textrm{struct}$\hspace{2pt} & {\color{red}2.9} & 4.7 & 8.7 & 13.0 & 18.6 & {\color{red}8.0} & 22.8 & 25.0 & 26.7 \\
\hline
\emph{All} &			ConSE~\cite{NorouziMBSSFCD14}\hspace{4pt} & 1.4 & 2.2 & 3.9 & 5.8 & 8.3 & 2.5 & 7.8 & 9.2 & 10.4 \\
 & ConSE by us & 1.3 & 2.1 & 3.8 & 5.8 & 8.7 & 3.2 & 9.2 & 10.7 & 12.0 \\ \cline{2-11}
						 & Ours$^\textrm{o-vs-o}$ & 1.4 & {\color{red}2.4} & {\color{red}4.5} & {\color{red}7.1} & {\color{red}10.9} & 3.1 & 9.0 & 10.9 & {\color{red}12.5} \\
						 & Ours$^\textrm{struct}$\hspace{2pt} & {\color{red}1.5} & {\color{red}2.4} & 4.4 & 6.7 & 10.0 & {\color{red}3.6} & {\color{red}9.6} & {\color{red}11.0} & 12.2 \\
\hline
\end{tabular}
\vskip -1em
\end{table*}

\subsubsection{Large-scale zero-shot learning}

One major limitation of most existing work on zero-shot learning is that the number of unseen classes is often small, dwarfed by the number of seen classes. However, real-world computer vision systems need to face a very large number of unseen objects. To this end, we evaluate our methods on the large-scale \textbf{ImageNet} dataset. 

Table~\ref{tbImagenet} summarizes our results and compares to the ConSE method~\cite{NorouziMBSSFCD14}, which is, to the best of our knowledge, the state-of-the-art method on this dataset.\footnote{We are aware of recent work by Lu~\cite{lu2015unsupervised} that introduces a novel form of semantic embeddings.} Note that  in some cases, our own implementation  (``ConSE by us'' in the table) performs slightly worse than the reported results, possibly attributed to differences in visual features, word vector embeddings, and other implementation details. Nonetheless, the proposed methods (using the same setting as ``ConSE by us'') always outperform both, especially in the very challenging scenario of \emph{All} where the number of unseen classes is 20,345, significantly larger than the number of seen classes. Note also that, for both types of metrics, when $K$ is larger, the improvement over the existing approaches is more pronounced.  It is also not surprising to notice that as the number of unseen classes increases from the setting \emph{2-hop} to \emph{All}, the performance of both our methods and ConSE degrade.

\subsubsection{Detailed analysis}

\begin{table*}
\centering
\small 
\caption{\small Detailed analysis of various methods: the effect of feature and attribute types on multi-way classification accuracies (in \%). Within each column, the best is in red and the 2nd best is in blue. We cite both previously published results (numbers in bold italics) and results  from our implementations of those competing methods (numbers in normal font) to enhance comparability and to ease analysis (see texts for details). We use the shallow features provided by~\cite{LampertNH14},~\cite{Jayaraman14},~\cite{PattersonH14} for \textbf{AwA}, \textbf{CUB}, \textbf{SUN}, respectively.}

\label{tbFeatures}
\small
\begin{tabular}{c|c|ccc|ccc}
\text{Methods}& Attribute & \multicolumn{3}{|c|}{Shallow features} & \multicolumn{3}{|c}{Deep features}\\ \cline{3-8}
&  type & \textbf{AwA} & \textbf{CUB} &   \textbf{SUN} & \textbf{AwA} & \textbf{CUB} & \textbf{SUN}\\ \hline
\text{DAP} \cite{LampertNH14}  & binary & \textbf{\emph{41.4} } & 28.3 & \textbf{\emph{22.2} } & \text{60.5 (50.0)} & \text{39.1 (34.8)} & \text{44.5} \\
\text{IAP} \cite{LampertNH14} & binary  &  \textbf{\emph{42.2} } & 24.4 & \textbf{\emph{18.0} } &   \text{57.2 (53.2)} & \text{36.7 (32.7)}  & \text{40.8} \\
\text{BN} \cite{WangJ13}  & binary & \textbf{\emph{43.4}}   & -  & - & - & - & -   \\
\text{ALE} \cite{AkataPHS13}$^\ddagger$ & binary & \textbf{\emph{37.4}} &\textbf{\emph{18.0}}$^\dagger$ & -&  -   & - &-   \\
\text{ALE}\hspace{17pt} & binary& \text{34.8} & \text{27.8}  & -&  \text{53.8 (48.8)} & \text{40.8 (35.3)} & 53.8 \\
\hline
\text{SJE} \cite{AkataRWLS15} & continuous  & \textbf{\emph{42.3}}$^\ddagger$ & \textbf{\emph{19.0}}$^\dagger$$^\ddagger$ & - &  \textbf{\emph{66.7 (61.9)}} &  \textbf{\emph{50.1 (40.3)}}$^\dagger$ & - \\
\text{SJE}\hspace{13pt} & continuous  & \text{36.2} & \text{34.6}  & - & \text{66.3 {(63.3)}} &\text{46.5 (42.8)} &\text{56.1}\\
\text{ESZSL} \cite{Bernardino15}$^\S$  & continuous  & \textbf{{\color{red}\emph{49.3}}}  & {\color{blue}37.0} & - & \text{59.6 (53.2)} & \text{44.0  (37.2)} & \text{8.7}\\
\text{ESZSL}\hspace{21pt} & continuous & {\color{blue}\text{44.1}}   & {\color{red}38.3} & - & \text{64.5 (59.4)} & \text{34.5 (28.0)} & \text{18.7}\\
\text{ConSE} \cite{NorouziMBSSFCD14}  & continuous & \text{36.5} & \text{23.7}  & -&  \text{63.3 (56.5)} & \text{36.2 (32.6)} & \text{51.9} \\
\text{COSTA} \cite{MensinkGS14}$^\sharp$ & continuous & 38.9 & 28.3 & - & 61.8 (55.2) & 40.8 (36.9) & 47.9 \\ 
\hline
{Ours$^\textrm{o-vs-o}$} & continuous & {42.6} & 35.0  & - & {\color{blue}69.7} {\color{blue}(64.0)} & {\color{blue}53.4 (46.6)} & {\color{red}62.8}\\
{Ours$^\textrm{cs}$}\hspace{10.5pt} & continuous & {42.1} & 34.7 & - & 68.4 {\color{red}(64.8)} & 51.6 (45.7) &  52.9\\
{Ours$^\textrm{struct}$}\hspace{2pt} & continuous & 41.5 & 36.4 & - & {\color{red}72.9} (62.8) &{\color{red}54.5 (47.1)} &  {\color{blue}62.7}\\
 \hline
\end{tabular}
\begin{flushleft}
\vskip -0.5em
$^\dagger$: Results reported by the authors on a particular seen-unseen split\eat{that is not publicly available}.\\
$^\ddagger$: Based on Fisher vectors as shallow features, different from those provided in~\cite{Jayaraman14,LampertNH14,PattersonH14}.\\
$^\S$: On the attribute vectors without $\ell_2$ normalization, while our own implementation shows that normalization helps in some cases.\\
$^\sharp$: As co-occurrence statistics are not available, we combine pre-trained classifiers with the weights defined in eq.~(\ref{simform}).
\end{flushleft}
\vskip -1em
\end{table*}

We experiment extensively to understand the benefits of many factors in our and other algorithms.  While trying all possible combinations is prohibitively expensive, we have provided a comprehensive set of results for comparison and drawing conclusions. %

\paragraph{Advantage of continuous attributes} It is clear from Table~\ref{tbFeatures} that, in general, continuous attributes as semantic embeddings for classes attain better performance than binary attributes. This is especially true when deep learning features are used to construct classifiers. It is somewhat expected that continuous attributes provide a more accurate real-valued similarity measure among classes. This presumably is exploited further by more powerful classifiers.

\paragraph{Advantage of deep features} It is also clear from Table~\ref{tbFeatures} that, across all methods, deep features significantly boost the performance based on shallow features. We use GoogLeNet and AlexNet (numbers in parentheses) and GoogLeNet generally outperforms AlexNet. 
It is worthwhile to point out that the reported results under deep features columns are obtained using linear classifiers, which outperform several nonlinear classifiers that use shallow features. This seems to suggest that deep features, often thought to be specifically adapted to seen training images, still work well when transferred to unseen images~\cite{FromeCSBDRM13}.

\begin{table}
\centering
\small 
\caption{\small Effect of types of semantic embeddings on \textbf{AwA}.}
\label{tSpace}
\small
\begin{tabular}{c|c|c}
\text{Semantic embeddings}&\text{Dimensions}& \text{Accuracy ($\%$)}
\\ \hline 
\text{word vectors} & 100   & 42.2 \\
\text{word vectors} & 1000   & 57.5 \\
\text{attributes}& 85   & 69.7\\
\text{attributes + word vectors} & 185 & 73.2 \\
\text{attributes + word vectors} & 1085 & \bf{76.3} \\
 \hline
\end{tabular}
\vskip -1em
\end{table}

\paragraph{Which types of semantic space?}  In Table~\ref{tSpace}, we show how effective our proposed method (Ours$^{\textrm{o-vs-o}}$) exploits the two types of semantic spaces: (continuous) attributes and word-vector embeddings on \textbf{AwA} (the only dataset with both embedding types). We find that attributes yield better performance than word-vector embeddings. However, combining the two gives the best result, suggesting that these two semantic spaces could be complementary and further investigation is ensured.

Table~\ref{tLearnSpace} takes a different view on identifying the best semantic space. We study whether we can learn optimally the semantic embeddings for the phantom classes that correspond to base classifiers. These preliminary studies seem to suggest that learning attributes could have a positive effect, though it is difficult to improve over word-vector embeddings.  We plan to study this issue more thoroughly in the future.
\begin{table}
\centering
\footnotesize
\caption{\small Effect of learning semantic embeddings}
\label{tLearnSpace}
\begin{tabular}{c|c|c|c}
Datasets & Types of embeddings & w/o learning & w/ learning\\ \hline
\textbf{AwA} & attributes & 69.7\% & 71.1\% \\ 
 & 100-d word vectors  & 42.2\% & 42.5\%\\
 & 1000-d word vectors & 57.6\% & 56.6\%\\ \hline
 \textbf{CUB} & attributes & 53.4\% & 54.2\%\\ \hline
 \textbf{SUN} & attributes & 62.8\% & 63.3\%\\ \hline
 \end{tabular}
 \vskip -1em
\end{table}

\begin{figure}
\centering
\small
\includegraphics[width=0.9\columnwidth]{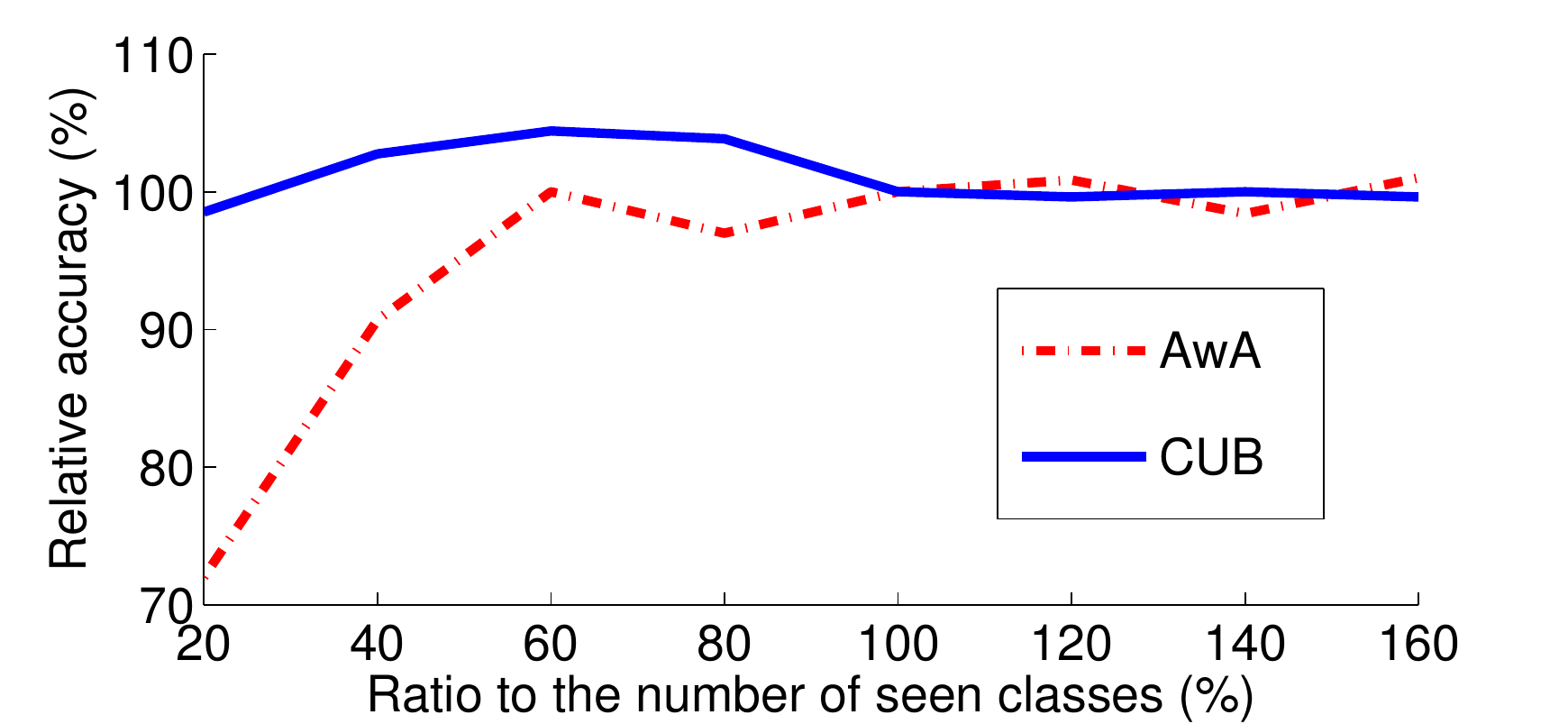}
\caption{\small We vary the number of phantom classes $\cR$ as a percentage of the number of seen classes $\cS$ and investigate how much that will affect classification accuracy (the vertical axis corresponds to the ratio with respect to the accuracy when $\cR = \cS$). The base classifiers are learned with Ours$^\textrm{o-vs-o}$.}
\label{fNumPhantom}
\vskip -1em
\end{figure}

\paragraph{How many base classifiers are necessary?} In Fig.~\ref{fNumPhantom}, we investigate how many base classifiers are needed --- so far, we have set that number to be the number of seen classes out of convenience. The plot shows that in fact, a smaller number (about 60\% -70\%) is enough for our algorithm to reach the plateau of the performance curve. Moreover, increasing the number of base classifiers does not seem to have an overwhelming effect. Further details and analysis are in the Suppl.

\section{Conclusion}
\label{sDiscuss}
We have developed a novel classifier synthesis mechanism for zero-shot learning by introducing the notion of ``phantom'' classes. The phantom classes connect the dots between the seen and unseen classes --- the classifiers of the seen and unseen classes are constructed from the same base classifiers for the phantom classes and with the same coefficient functions. As a result, we can conveniently learn the classifier synthesis mechanism leveraging labeled data of the seen classes and then readily apply it to the unseen classes. Our approach outperforms the state-of-the-art methods on four benchmark datasets in most scenarios.
\section*{Acknowledgments}
B.G. is partially supported by NSF IIS-1566511. Others are partially supported by USC\eat{Provost and} Annenberg Graduate Fellowship, NSF IIS-1065243, 1451412, 1513966, 1208500, CCF-1139148, a Google Research Award, an Alfred. P. Sloan Research Fellowship and ARO\# W911NF-12-1-0241 and W911NF-15-1-0484.

{\small
\bibliographystyle{ieee}
\bibliography{main_cvpr}
}

\clearpage

{
\vskip .375in
\begin{center}
{\Large \bf Supplementary Material:\\Synthesized Classifiers for Zero-Shot Learning \par}
\vspace*{24pt}
\end{center}
}
\appendix
In this Supplementary Material, we provide details omitted in the main text.
\begin{itemize}
\item Section~\ref{sCV}: cross-validation strategies (Section~3.2 of the main paper).
\item Section~\ref{sMetric}: learning metrics for semantic similarity (Section~3.1 of the main paper).
\item Section~\ref{det_exp_setup}: details on experimental setup (Section~4.1 of the main paper).
\item Section~\ref{det_imp}: implementation details (Section~4.1 and 4.2.3 of the main paper).
\item Section~\ref{sExtraExp}: additional experimental results and analyses (Section~4.2 of the main paper).
\end{itemize}

\section{Cross-validation (CV) strategies}
\label{sCV}
\begin{figure*}[ht]
\centering
\vspace{-.1in}
\includegraphics[width=0.65\textwidth]{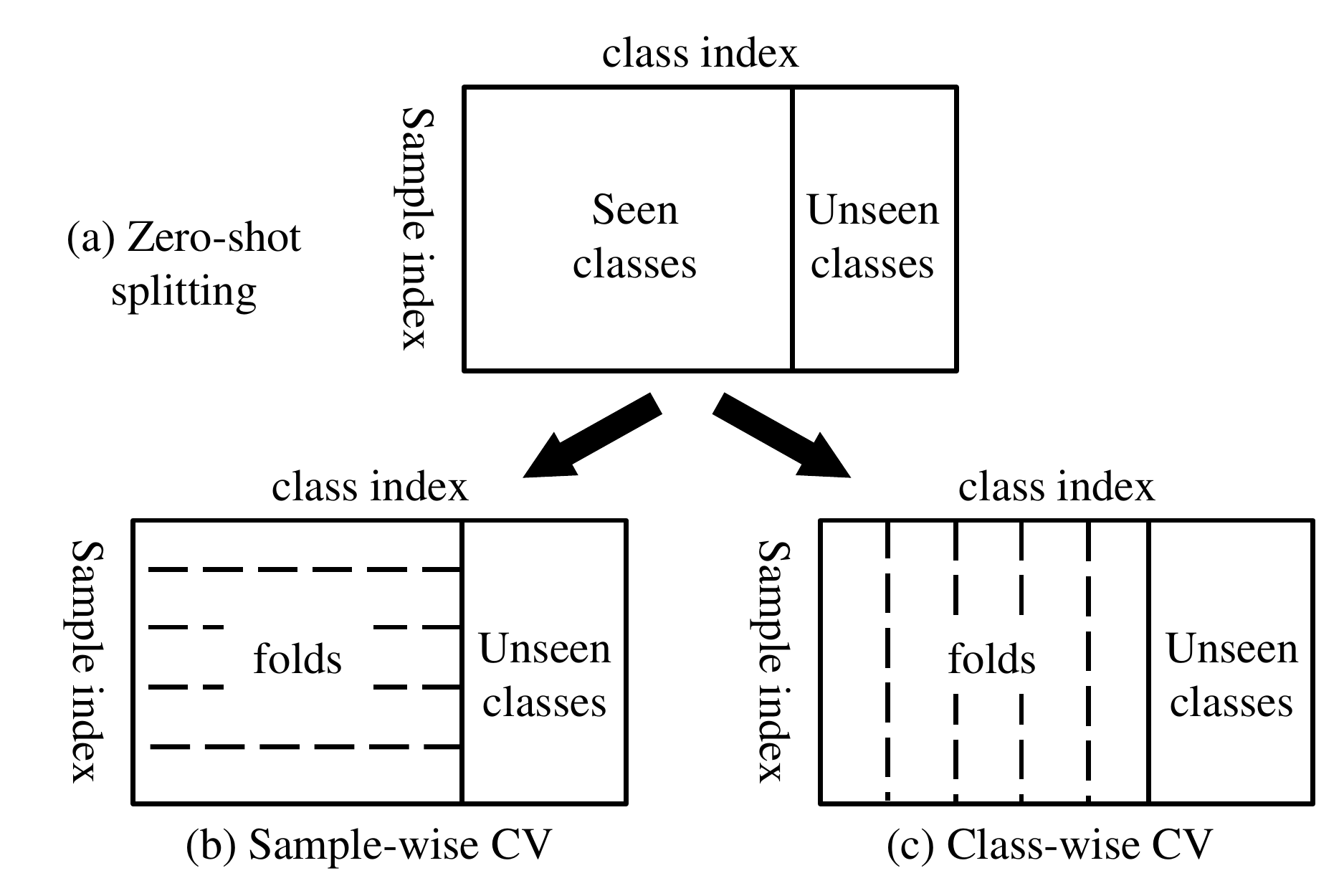}
\vspace{-.1in}
\caption{\small Data splitting for different cross-validation (CV) strategies: (a) the seen-unseen class splitting for zero-shot learning, (b) the sample-wise CV, (c) the class-wise CV (cf. Section 3.2 of the main paper).}
\label{fig:1}
\end{figure*} 
There are a few free hyper-parameters in our approach (cf.\ Section~3.2 of the main text). To choose the hyper-parameters in the conventional cross-validation (CV) for multi-way classification, one splits the training data into several folds such that they share the same set of class labels with one another. Clearly, this strategy is not sensible for zero-shot learning as it does not imitate what actually happens at the test stage.
We thus introduce a new strategy for performing CV, inspired by the hyper-parameter tuning in~\cite{Bernardino15}. The key difference of the new scheme to the conventional CV is that we split the data into several folds such that the class labels of these folds are disjoint. For clarity, we denote the conventional CV as \emph{sample}-wise CV and our scheme as \emph{class}-wise CV. Figure \ref{fig:1}(b) and \ref{fig:1}(c) illustrate the two scenarios, respectively. We empirically compare them in Section~\ref{sCVexp}. Note that several existing models~\cite{AkataRWLS15,ElhoseinySE13,Bernardino15,zhang2015zero} also follow similar hyper-prameter tuning procedures.
\subsection{Learning semantic embeddings}
We propose an optimization problem for learning semantic embeddings in Section~3.2 of the main text. There are four hyper-parameters $\lambda, \sigma, \eta,$ and $\gamma$ to be tuned. To reduce the search space during cross-validation, we first fix $\vb_r = \va _r$ for $r = 1,\ldots,\cR$ and tune $\lambda, \sigma$. Then we fix $\lambda$ and $\sigma$ and tune $\eta$ and $\gamma$.

\section{Learning metrics for computing similarities between semantic embeddings}
\label{sMetric}
Recall that, in Section~3.1 of the main text, the weights in the bipartite graph are defined based on the distance $d(\va_c,\vb_r)=(\va_c-\vb_r)^T\mSigma^{-1}(\va_c-\vb_r)$. In this section, we describe an objective for learning a more general Mahalnobis metric than $\Sigma^{-1} = \sigma^2 \mI$. We focus on the case when $\cR = \cS$ and on learning a diagonal metric $\Sigma^{-1} = \mM^T\mM$, where $\mM$ is also diagonal. We solve the following optimization problem. 
\begin{align}
&\min_{\mM, \vv_1,\cdots,\vv_\cR} \sum_{c=1}^{\cS}\sum_{n=1}^\cN \ell({\vx_n}, \I_{y_n,c}; {\vw_c}) \\
&+ \frac{\lambda}{2} \sum_{r=1}^{\cR} \twonorm{\vv_r} + \frac{\gamma}{2} \left\|\mM - \sigma\mI\right\|_F^2, \\
&\mathsf{s.t.}\quad \vw_c = \sum_{r=1}^\cR s_{cr}\vv_r, \quad \forall\, c\in\mathcal{T}=\{1,\cdots,\cS\}  \notag
\end{align}
where $\ell(\vx, y; \vw)=\max(0,1-y\vw\T\vx)^2$ is the squared hinge loss. The indicator $\I_{y_n,c}\in\{-1,1\}$ denotes whether or not $y_n=c$.

Again, we perform alternating optimization for minimizing the above objective function. At first, we fix $\mM = \sigma\mI$ and optimize $\{\vv_1,\cdots,\vv_\cR\}$ to obtain a reasonable initialization. Then we perform alternating optimization. To further prevent over-fitting, we alternately optimize $\mM$ and $\{\vv_1,\cdots,\vv_\cR\}$ on different but overlapping subsets of training data. In particular, we split data into 5 folds and optimize $\{\vv_1,\cdots,\vv_\cR\}$ on the first 4 folds and $\mM$ on the last 4 folds. We report results in Section~\ref{sMetricExp}.

\section{Details on the experimental setup}
\label{det_exp_setup}
We present details on the experimental setup in this section and additional results in Section~\ref{sExtraExp}.

\subsection{Datasets}

We use four benchmark datasets in our experiments. The \textbf{Animals with Attributes (AwA)} dataset~\cite{LampertNH14} consists of 30,475 images of 50 animal classes. Along with the dataset, a standard data split is released for zero-shot learning: 40 seen classes (for training) and 10 unseen classes. 
The second dataset is the \textbf{CUB-200-2011 Birds (CUB)}~\cite{WahCUB_200_2011}. It has 200 bird classes and 11,788 images.  We randomly split the 200 classes into 4 disjoint sets (each with 50 classes) and treat each of them as the unseen classes in turn. We report the average results from the four splits. 
The \textbf{SUN Attribute (SUN)} dataset~\cite{PattersonH14} contains 14,340 images of 717 scene categories (20 images from each category). \eat{The dataset is drawn from the the \textbf{SUN} database~\cite{XiaoHEOT10}.} Following ~\cite{LampertNH14}, we randomly split the 717 classes into 10 disjoint sets (each with 71 or 72 classes) in a similar manner to the class splitting on \text{CUB}. We note that some previous published results~\cite{JayaramanG14,Bernardino15,zhang2015classifying,zhang2015zero} are based on a simpler setting with 707 seen and 10 unseen classes. For comprehensive experimental comparison, we also report our results on this setting in Table~\ref{tbOtherMethods}.

For the large-scale zero-shot experiment on the \textbf{ImageNet} dataset~\cite{deng2009imagenet}, we follow the setting in \cite{FromeCSBDRM13, NorouziMBSSFCD14}. The ILSVRC 2012 1K dataset~\cite{ILSVRC15} contains 1,281,167 training and 50,000 validation images from 1,000 categories and  is treated as the seen-class data. Images of unseen classes come from the rest of the ImageNet Fall 2011 release dataset~\cite{deng2009imagenet} that do not overlap with any of the 1,000 categories. We will call this release the ImageNet 2011 21K dataset (as in \cite{FromeCSBDRM13, NorouziMBSSFCD14}). Overall, this dataset contains 14,197,122 images from 21,841 classes, and we conduct our experiment on \textbf{20,842 unseen classes}\footnote{There is one class in the ILSVRC 2012 1K dataset that does not appear in the ImageNet 2011 21K dataset. Thus, we have a total of 20,842 unseen classes to evaluate.}.

\subsection{Semantic spaces}
\paragraph{SUN}
Each image is annotated with 102 continuous-valued attributes. For each class, we average attribute vectors over all images belonging to that class to obtain a class-level attribute vector.

\paragraph{ImageNet}
We train a skip-gram language model~\cite{MikolovCCD13, MikolovSCCD13} on the latest Wikipedia dump corpus\footnote{\url{http://dumps.wikimedia.org/enwiki/latest/enwiki-latest-pages-articles.xml.bz2} on September 1st, 2015} (with more than 3 billion words) to extract a 500-dimensional word vector for each class. In particular, we train the model using the word2vec package\footnote{\url{https://code.google.com/p/word2vec/}} --- we preprocess the corpus with the word2phrase function so that we can directly obtain word vectors for both single-word and multiple-word terms, including those terms in the ImageNet \emph{synsets}\footnote{Each class of \textbf{ImageNet} is a \emph{synset}: a set of synonymous terms, where each term is a word or phrase.}.
We impose no restriction on the vocabulary size. Following \cite{FromeCSBDRM13}, we use a window size of 20, apply the hierarchical softmax for predicting adjacent terms, and train the model with a single pass through the corpus. As one class may correspond to multiple word vectors by the nature of synsets, we simply average them to form a single word vector for each class. We obtain word vectors for all the 1,000 seen classes and for \textbf{20,345} (out of 20,842) unseen classes. We ignore classes without word vectors in the experiments.

\subsection{Visual features}

We denote features that are not extracted by deep learning as shallow features.

\paragraph{Shallow features}
On \textbf{AwA}, many existing approaches take traditional features such as color histograms, SIFT, and PHOG that come with the dataset~\cite{LampertNH14,RohrbachSSGS10,WangJ13}, while others use the Fisher vectors~\cite{AkataPHS13,AkataRWLS15}. 
The \textbf{SUN} dataset also comes with several traditional shallow features, which are used in~\cite{JayaramanG14,LampertNH14,Bernardino15}. 
\eat{For \textbf{CUB} and \textbf{ImageNet}, no features are released.}

In our experiments, we use the shallow features provided by~\cite{LampertNH14} ,~\cite{Jayaraman14}, and~\cite{PattersonH14} for \textbf{AwA}, \textbf{CUB}, and \textbf{SUN}, respectively, unless stated otherwise. 

\paragraph{Deep features} Given the recent impressive success of deep Convolutional Neural Networks (CNNs)~\cite{KrizhevskySH12} on image classification, we conduct experiments with deep features on all datasets. We use the Caffe package~\cite{jia2014caffe} to extract AlexNet~\cite{KrizhevskySH12} and GoogLeNet~\cite{SzegedyLJSRAEVR14} features for images from \textbf{AwA} and \textbf{CUB}. Observing that GoogLeNet give superior results over AlexNet on \textbf{AwA} and \textbf{CUB}, we focus on GoogLeNet features on large datasets: \textbf{SUN} and \textbf{ImageNet}. These networks are pre-trained on the ILSVRC 2012 1K dataset~\cite{deng2009imagenet, ILSVRC15} for \textbf{AwA}, \textbf{CUB}, and \textbf{ImageNet}. For \textbf{SUN}, the networks are pre-trained on the Places database~\cite{ZhouLXTO14}, which was shown to outperform the networks pre-trained on ImageNet on scene classification tasks. For AlexNet, we use the 4,096-dimensional activations of the penultimate layer (fc7) as features, and for GoogLeNet we extract features by the 1,024-dimensional activations of the pooling units following the suggestion by~\cite{AkataRWLS15}.

For \textbf{CUB}, we crop all images with the provided bounding boxes. For \textbf{ImageNet}, we center-crop all images and do not perform any data augmentation or other preprocessing.

\subsection{Evaluation protocols on ImageNet}
When computing Hierarchical precision@K (HP@K), we use the algorithm in the Appendix of \cite{FromeCSBDRM13} to compute a set of at least $K$ classes that are considered to be correct. This set is called $hCorrectSet$ and it is computed for each $K$ and class $c$. See Algorithm~\ref{aHCorrect} for more details. The main idea is to expand the radius around the true class $c$ until the set has at least $K$ classes.
\vspace{.1in}
\begin{algorithm}
\small{
\caption{Algorithm for computing $hCorrectSet$ for $H@K$~\cite{FromeCSBDRM13}}
\begin{algorithmic}[1]
	\State Input: $K$, class $c$, ImageNet hierarchy
	\State $hCorrectSet \leftarrow \emptyset$
	\State $R \leftarrow 0$
	\While {NumberElements($hCorrectSet) < K$}
		\State $radiusSet \leftarrow$ all nodes in the hierarchy which are $R$ hops from $c$
		\State $validRadiusSet \leftarrow$ ValidLabelNodes($radiusSet$)
		\State $hCorrectSet \leftarrow$ $hCorrectSet \cup validRadiusSet$
		\State $R \leftarrow R + 1$
	\EndWhile
	\vspace{0.1in}\State \Return $hCorrectSet$
\end{algorithmic}
\label{aHCorrect}
}
\end{algorithm}

Note that $validRadiusSet$ depends on which classes are in the label space to be predicted (i.e., depending on whether we consider \emph{2-hop}, \emph{3-hop}, or \emph{All}. We obtain the label sets for \emph{2-hop} and \emph{3-hop} from the authors of \cite{FromeCSBDRM13, NorouziMBSSFCD14}. We implement Algorithm~\ref{aHCorrect} to derive $hCorrectSet$ ourselves. 

\section{Implementation details}
\label{det_imp}
\subsection{How to avoid over-fitting?}
Since during training we have access only to data from the seen classes, it is important to avoid over-fitting to those seen classes. We apply the \emph{class-wise} cross-validation strategy (Section~\ref{sCV}), and restrict the semantic embeddings of phantom classes to  be close to the semantic embeddings of seen classes (Section 3.2 of the main text).

\subsection{Combination of attributes and word vectors}
In Table 5 of the main text and Section~\ref{sEmbed} of this material, we combine attributes and word vectors to improve the semantic embeddings.  We do so by first computing $s_{rc}$ in eq. (2) of the main text for different semantic sources, and then perform convex combination on $s_{rc}$ of different sources to obtain the final $s_{rc}$. The combining weights are determined via cross-validation.\\

\subsection{Initialization}
All variables are randomly initialized, unless stated otherwise. Other details on initialization can be found in Section 3.2 of the main text and Section~\ref{sMetric} and \ref{sVaryBase} of this material. 

\subsection{ConSE~\cite{NorouziMBSSFCD14}}
Instead of using the CNN 1K-class classifiers directly, we train (regularized) logistic regression classifiers using recently released multi-core version of LIBLINEAR \cite{fan2008liblinear}.
Furthermore, in \cite{NorouziMBSSFCD14}, the authors use the averaged word vectors for seen classes, but keep for each unseen class the word vectors of all synonyms. In other words, each unseen class can be represented by multiple word vectors. In our implementation, we use averaged word vectors for both seen and unseen classes for fair comparison.

\subsection{Varying the number of base classifiers}
\label{sVaryBase}

In Section 4.2.3 and Figure 2 of the main text, we examine the use of different numbers of base classifiers (i.e., $\cR$). The semantic embedding $\vb_r$ of the phantom classes are set equal to $\va_r, \forall r\in\{1, \cdots, \cR\}$ at 100\% (i.e., $\cR=\cS$). For percentages smaller than 100\%, we perform $K$-means and set $\vb_r$ to be the cluster centroids after $\ell_2$ normalization (in this case, $\cR = K$). For percentages larger than 100\%, we set the first $\cS$ $\vb_r$ to be $\va_r$, and the remaining $\vb_r$ as the random combinations of $\va_r$ (also with $\ell_2$ normalization on $\vb_r$).
\section{Additional experimental results and analyses}
\label{sExtraExp}

We present in this section some additional experimental results on zero-shot learning. \emph{Unless stated otherwise, we focus on learning with the one-versus-other loss (cf. eq. (5) of the main text).} 

\subsection{Cross-validation (CV) strategies}  \label{sCVexp}
Table~\ref{tTwoCVs} shows the results on \textbf{CUB} (averaged over four splits) using the hyper-parameters tuned by class-wise CV and sample-wise CV, respectively. The results based on class-wise CV are about 2\% better than those of sample-wise CV, verifying the necessity of simulating the zero-shot learning scenario while we tune the hyper-parameters at the training stage.

\begin{table}
\centering
\small
\vspace{0.2in}
\caption{\small Comparison between sample- and class-wise cross-validation for hyper-parameter tuning on \textbf{CUB} (learning with the one-versus-other loss).}
\label{tOVP}
\small
\begin{tabular}{c|c|c}

\text{CV}&\textbf{CUB}&\textbf{CUB}\\
\text{Scenarios}&\text{(AlexNet)}&\text{(GoogLeNet)}
\\ \hline
\text{Sample-wise}& 44.7 & 52.0\\
\text{Class-wise}& 46.6 & 53.4 \\ \hline
\end{tabular}
\vskip 1em
\label{tTwoCVs} 
\end{table}

\subsection{Additional comparison of different semantic spaces for embedding classes}
\label{sEmbed}
Our method for synthesizing classifiers accepts different semantic embedding spaces. We expand our results on Table 5 of the main text to include AlexNet features as well. The results are in Table~\ref{tWordAtt}. We use word vectors provided by Fu et al.~\cite{FuHXFG14,fu2015pami}, which are of 100 and 1000 dimensions per class, respectively. We see that the two types of features, AlexNet and GoogLeNet, demonstrate very similar trends. First, higher-dimensional word vectors often give rise to better performance. Second, human-annotated attributes outperform automatically-learned word vectors. Third, combining the word vectors and the attributes leads to better results than separately using either one of them.
\eat{Attributes, as defined and annotated by human experts, outperform the unsupervisedly learned word vectors by over 10\% of accuracy. Thus we will use attributes in our remaining experiments. \WC{We further include our results using attributes + word vectors together as the semantic information.}}

\begin{table*}
\centering
\small 
\caption{\small Comparison between different semantic embedding spaces for our approach.}
\vskip 1em
\small
\begin{tabular}{c|c|c|c|c}
\text{Methods}&\text{Semantic embeddings}&\text{Dimensions}&\text{Features}&\textbf{AwA}
\\ \hline 
\text{Ours$^\textrm{o-vs-o}$}&\text{word vectors}& 100&\text{AlexNet} & 37.6\\
\text{Ours$^\textrm{o-vs-o}$}&\text{word vectors}& 1000&\text{AlexNet} & 52.4\\
\text{Ours$^\textrm{o-vs-o}$}&\text{attributes} & 85 &\text{AlexNet} & 64.0\\
\text{Ours$^\textrm{o-vs-o}$}&\text{attributes + word vectors} & 85 + 100 &\text{AlexNet} & 65.6 \\
\text{Ours$^\textrm{o-vs-o}$}&\text{attributes + word vectors} & 85 + 1000 &\text{AlexNet} & \bf{68.0} \\
\hline
\text{SJE\cite{AkataRWLS15}}&\text{word vectors} & 400 &\text{GoogLeNet} & 51.2 \\
\text{Ours$^\textrm{o-vs-o}$}&\text{word vectors} & 100 &\text{GoogLeNet} & 42.2 \\
\text{Ours$^\textrm{o-vs-o}$}&\text{word vectors} & 1000 &\text{GoogLeNet} & 57.5 \\
\text{Ours$^\textrm{o-vs-o}$}&\text{attributes}& 85 &\text{GoogLeNet} & 69.7\\
\text{Ours$^\textrm{o-vs-o}$}&\text{attributes + word vectors} & 85 + 100 &\text{GoogLeNet} & 73.2 \\
\text{Ours$^\textrm{o-vs-o}$}&\text{attributes + word vectors} & 85 + 1000 &\text{GoogLeNet} & \bf{76.3} \\
 \hline
\end{tabular}
\vskip 1em
\label{tWordAtt}
\vskip 1em
\end{table*} 
 
\subsection{Comparison to other state-of-the-art methods}
\label{sOthermethods}
In Table~\ref{tbOtherMethods}, we contrast our methods to several other state-of-the-art methods, in addition to Table 2 of the main text. \eat{In future work, we will aim to replicate their experiment settings so that a fair comparison can be conducted.} We note subtle differences in the experiment setup of some of these methods from ours:
\begin{itemize}
\item \text{TMV-BLP and TMV-HLP}~\cite{FuHXFG14,fu2015pami}. These methods focus on the transductive setting, where they have access to unlabeled test data from unseen classes during the training stage. Additionally, they use OverFeat~\cite{sermanet2014overfeat} features for \textbf{CUB}, OverFeat+DeCAF~\cite{donahue2013decaf} for \textbf{AwA}, and both attributes and word vectors for class embeddings.
\item \cite{Kodirov_2015_ICCV}. This method works on the transductive setting. It uses OverFeat features for both \textbf{AwA} and \textbf{CUB}, and combines attributes and word vectors for class embeddings.
\item \cite{Li_2015_ICCV} This method works on the semi-supervised setting, where a portion of unlabeled data (not used for testing) from unseen classes are available at training.
\item \text{HAT-n} ~\cite{AlHalahS15}. This method uses extra semantic information (WordNet class hiearchy).  It uses CNN-M2K features~\cite{Chatfield14} and extra cropped images.
\item \text{AMP (SR+SE)}~\cite{fu2015zero}. This method uses attributes and (100-dimensional) word vectors and OverFeat features in their experiments.
\item~\cite{YuCFSC13}. This method focuses on mining/discovering new (category-level) attributes. It requires extra human efforts to annotate the new attributes for unseen classes.
 \item~\cite{JayaramanG14}. The best result on \textbf{AwA} presented in this paper uses the discovered attributes in~\cite{YuCFSC13}.
\end{itemize}
As shown in Table~\ref{tbOtherMethods}, our method outperforms all of them on the dataset \textbf{CUB} despite the fact that they employ extra images or semantic embedding information.

\subsubsection{\textbf{SUN-10}} Some existing work~\cite{JayaramanG14,Bernardino15,zhang2015classifying,zhang2015zero} considers another setting for \textbf{SUN} dataset ---  with 707 seen classes and 10 unseen classes. Moreover, \cite{Bernardino15,zhang2015classifying,zhang2015zero} use the VGG-verydeep-19~\cite{Simonyan15c} CNN features. In Table~\ref{tbOtherMethods}, we provide results of our approach based on this splitting. Compared to previously published results, our method again clearly shows superior performance.  

\begin{table*}
\centering
\small 
\caption{\small Comparison between our method and other state-of-the-art methods for zero-shot learning. Our results are based on the GoogLeNet features and attributes or attributes + word vectors as semantic embeddings. See Section~\ref{sOthermethods} for the details of other methods and~\textbf{SUN-10}.} \label{tbOtherMethods}
\vskip 0.25em
\small
\begin{tabular}{c|ccc|cccc}
\text{Methods}& \multicolumn{3}{|c|}{Shallow features} & \multicolumn{4}{|c}{Deep features}\\ \cline{2-8}
& \textbf{AwA} & \textbf{CUB} & \textbf{SUN-10} & \textbf{AwA} & \textbf{CUB} & \textbf{SUN} & \textbf{SUN-10} \\ \hline
\text{TMV-BLP} \cite{FuHXFG14} & \textbf{\emph{47.7}}  & \textbf{\emph{16.3}}$^\dagger$$^\ddagger$ & - & {\color{blue}\textbf{\emph{77.8}}} & \textbf{\emph{45.2}}$^\dagger$ & - & -\\
\text{TMV-HLP} \cite{fu2015pami} & {\color{blue}\textbf{\emph{49.0}}}  & \textbf{\emph{19.5}}$^\dagger$$^\ddagger$ & - & {\color{red}\textbf{{\color{red}\emph{80.5}}}} & \textbf{\emph{47.9}}$^\dagger$ & - & -\\
\cite{Kodirov_2015_ICCV} & {\color{red}\textbf{\emph{49.7}}} & {\textbf{\emph{28.1}}}$^\dagger$$^\ddagger$ & - & {\textbf{\emph{75.6}}} & {\textbf{\emph{40.6}}}$^\dagger$ & - & - \\
\cite{Li_2015_ICCV} & \textbf{\emph{40.0}} & - & - & - & - & - & - \\
\text{HAT-n}  \cite{AlHalahS15} & - & - & - & \textbf{\emph{68.8}} & \textbf{\emph{48.6}}$^\dagger$& - & - \\
\text{AMP (SR+SE)} \cite{fu2015zero} & - & - & - & \textbf{\emph{66.0}} & - & - & - \\
\cite{YuCFSC13} & \textbf{\emph{48.3}} & - & - & - & - & - & - \\
\cite{JayaramanG14} & {\textbf{\emph{48.7}}} & - & {\color{blue}56.2} & - & - & - & - \\
\text{ESZSL}~\cite{Bernardino15} & - & - & {\color{red}65.8} & - & - & - & \textbf{\emph{82.1}} \\
\text{SSE-ReLU}~\cite{zhang2015zero} & - & - & - & - & - & - & \textbf{\emph{82.5}} \\
\cite{zhang2015classifying} & - & - & - & - & - & - & \textbf{\emph{83.8}} \\
\text{SJE}\cite{AkataRWLS15} & - & - & - & - & - & - & {\color{blue}{87.0}} \\
\hline
{Ours$^\textrm{o-vs-o}$}  & {42.6} & {\color{blue}35.0}  & - & {69.7} & {\color{blue}53.4} & {\color{red}62.8} & {\color{red} 90.0}\\
{Ours$^\textrm{cs}$}  & {42.1} & 34.7 & - & {68.4} & 51.6 &  52.9 & {\color{blue}87.0} \\
{Ours$^\textrm{struct}$} & 41.5 & {\color{red}36.4} & - & 72.9 &{\color{red}54.5} &  {\color{blue}62.7} & 85.0 \\
\hline
$^\S${Ours$^\textrm{o-vs-o}$}  & - & -  & - & 76.3 & - & - & - \\
\hline
\end{tabular}
\vskip 0.25em
\begin{flushleft}
$^\dagger$: Results reported by the authors on a particular seen-unseen split\eat{that is not publicly available}.\\
$^\ddagger$: Based on Fisher vectors as shallow features.\\
$^\S$: Ours with attributes + (1000-dimensional) word vectors.\\
\end{flushleft}
\end{table*}

\subsection{Discussion on the numbers of seen and unseen classes}
In this subsection, we analyze the results under different numbers of seen/unseen classes in performing zero-shot learning using the \textbf{CUB} dataset. 

\subsubsection{Varying the number of seen classes} We first examine the performance of zero-shot learning under different numbers of seen classes (e.g., 50, 100, and 150) while fixing the number of unseen classes to be 50. We perform 20 random selections of seen/unseen classes. Unsurprisingly, Table~\ref{tb1_diff_seen} shows that increasing the number of seen classes in training leads to improved performance on zero-shot learning.

\begin{table}[t]
\centering
\small 
\caption{\small Performance of our method under different number $\cS$ of seen classes on \textbf{CUB}. The number of unseen classes $\cU$ is fixed to be 50.}
\vskip 1em
\small
\begin{tabular}{c|c|c|c}
$\cU\text{ and }\cS$ & $\cS=50$ & $\cS=100$ & $\cS=150$ \\ \hline 
$\cU=50$& 38.4& 49.9 & 53.8\\ \hline 
\end{tabular}
\vskip 1em
\label{tb1_diff_seen}
\vskip 1em
\end{table}  

\subsubsection{Varying the number of unseen classes} 
\begin{figure}
\centering
\includegraphics[width=0.48\textwidth]{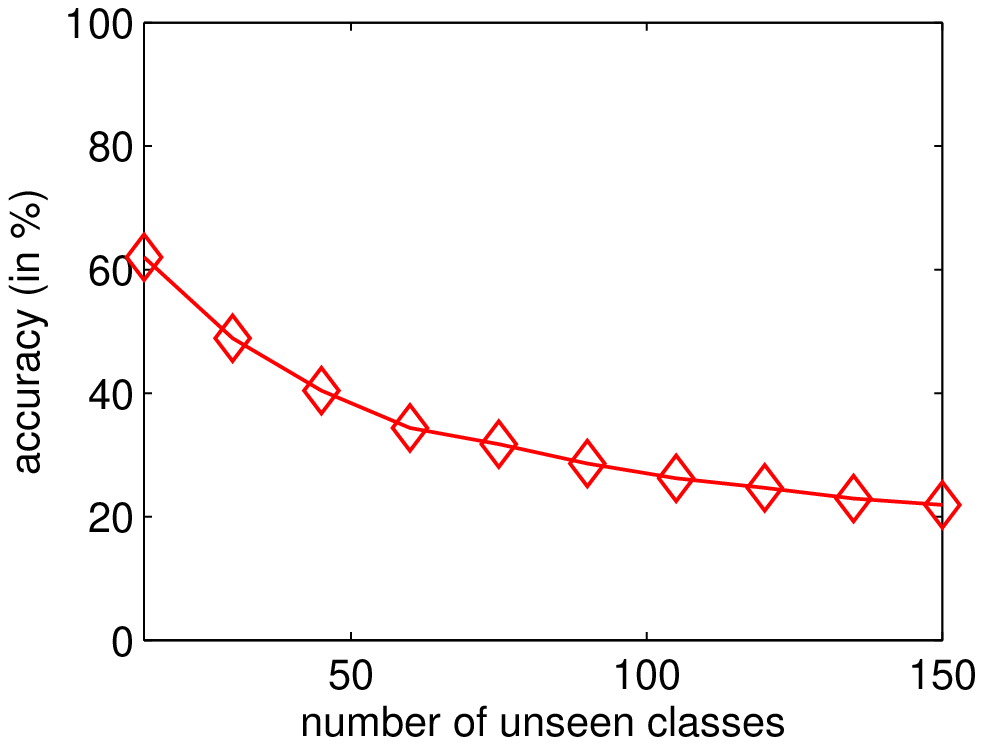}
\caption{\small Performance of our method under different numbers of unseen classes on \textbf{CUB}. The number of seen classes is fixed to be 50.}
\label{fig:3}
\end{figure}

We then examine the performance of our approach to zero-shot learning under different numbers of unseen classes (e.g., within [0, 150]), with the number of seen classes fixed to be 50 during training. We again conduct experiments on \textbf{CUB}, and perform 20 random selections of seen/unseen classes. The results are presented in Figure~\ref{fig:3}. We see that the accuracy drops as the number of unseen classes increases.

\subsection{Results on learning metrics for computing semantic similarities}
\label{sMetricExp}
We improve our method by also learning metrics for computing semantic similarity. Please see Section~\ref{sMetric} for more details.
Preliminary results on \textbf{AwA} in Table~\ref{tMetric} suggest that learning metrics can further improve upon our current one-vs-other formulation. 

\begin{table}
\centering
\small
\caption{\small Effect of learning metrics for computing semantic similarity on \textbf{AwA}.}
\label{tMetric}
\begin{tabular}{c|c|c|c}
Dataset & Type of embeddings & w/o learning & w/ learning\\ \hline
\textbf{AwA} & attributes & 69.7\% & 73.4\% \\ \hline
 \end{tabular}
 \vskip -1em
\end{table}

\subsection{Detailed results and analysis of experiments on ImageNet}
\label{det_ImageNet}
Table~\ref{tbImagenetSup} provides expanded zero-shot learning results on \textbf{ImageNet} (cf. Table 3 of the main text). Note that ConSE~\cite{NorouziMBSSFCD14} has a free parameter $T$, corresponding to how many nearest seen classes to use for convex combination. In our implementation, we follow the paper to test on $T=$ 1, 10, and 1,000. We further apply the class-wise cross-validation (cf. Section \ref{sCV} of this material) to automatically set $T$. We also report the published best result in~\cite{NorouziMBSSFCD14}. Our methods (Ours$^\textrm{o-vs-o}$ and Ours$^\textrm{struct}$) achieve the highest accuracy in most cases.

\begin{figure*}
\centering
\small
\includegraphics[width=0.9\textwidth]{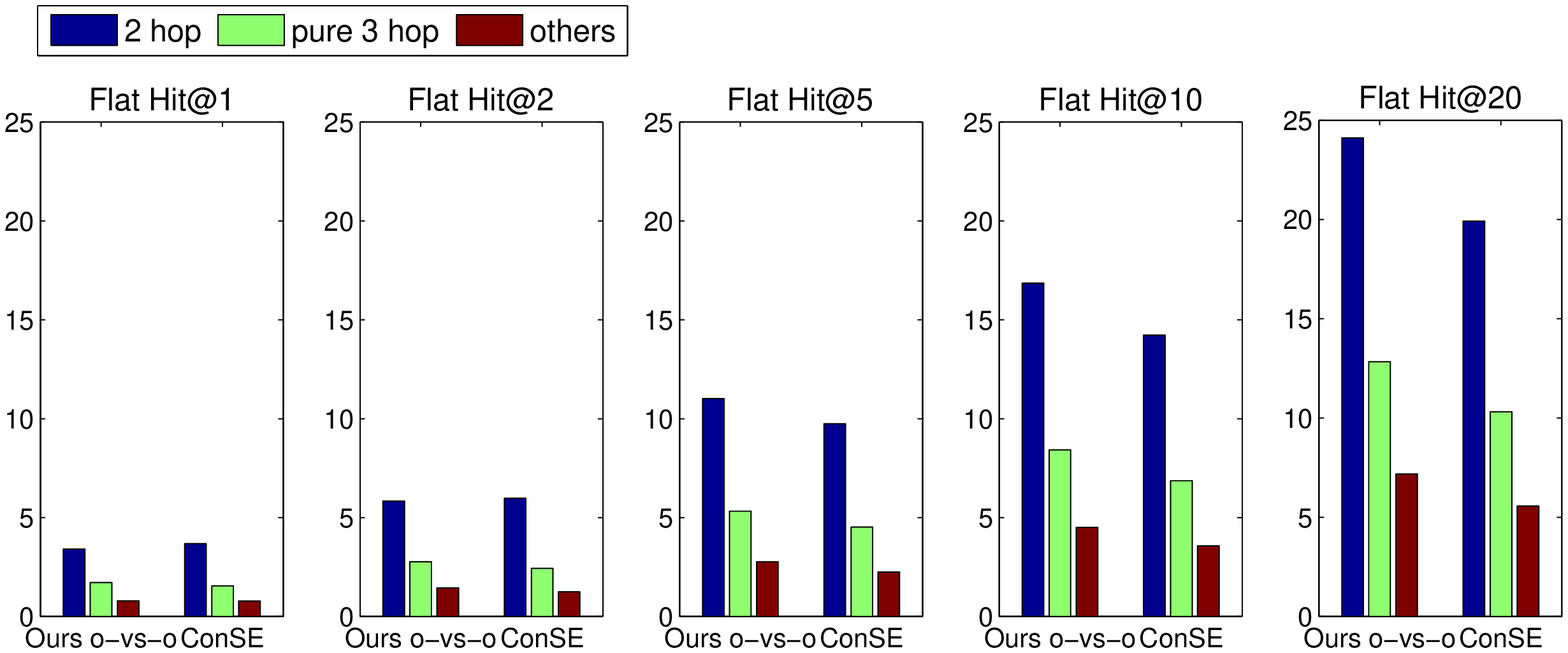}
\vspace{-2pt}
\caption{\small On the \textbf{ImageNet} dataset, we outperform ConSE (i.e., ConSE(10) of our implementation) on different \emph{disjoint} sets of categories in the scenario \emph{All} in almost all cases. See Section~\ref{det_ImageNet} of this material for details.}
\label{fImagenet}
\vspace{-5pt}
\end{figure*}

As mentioned in the main text, the three sets of unseen classes, \emph{2-hop}, \emph{3-hop}, and \emph{All} are built according to the ImageNet label hierarchy. Note that they are not mutually exclusive. Indeed, \emph{3-hop} contains all the classes in \emph{2-hop}, and \emph{All} contains all in \emph{3-hop}. To examine if the semantic similarity/dissimilarity to the 1K seen classes (according to the label hierarchy) would affect the classification accuracy, we split \emph{All} into three \emph{disjoint} sets, \emph{2-hop}, \emph{pure 3-hop}, and \emph{others}, which contain 1,509, 6,169, and 12,667 classes, respectively (totally 20,345). We then test on \emph{All}, but report accuracies of images belonging to different \emph{disjoint} sets separately. Figure~\ref{fImagenet} summarizes the results. Our method outperforms ConSE in almost all cases. The decreasing accuracies from \emph{2-hop}, \emph{pure 3-hop}, to \emph{others} (by both methods) verify the high correlation of the semantic relationship to the classification accuracy in zero-shot learning. This observation suggests an obvious potential limitation: it is unrealistic to expect good performance on unseen classes that are semantically too dissimilar to seen classes.

\begin{table*}
\centering
\small 
\caption{\small Flat Hit@K and Hierarchial precision@K performance (in \%) on the task of zero-shot learning on \textbf{ImageNet}. 
We mainly compare it with ConSE($T$) \cite{NorouziMBSSFCD14}, where $T$ is the number of classifiers to be combined in their paper. For ConSE(CV), $T$ is obtained by class-wise CV. Lastly, the best published results in \cite{NorouziMBSSFCD14} are also reported, corresponding to ConSE(10) \cite{NorouziMBSSFCD14}. For both types of metrics, the higher the better.} 
\label{tbImagenetSup}
\vskip 0.25em
\small
\begin{tabular}{c|c|ccccc|cccc}
\text{Scenarios} & \text{Methods} & \multicolumn{5}{|c|}{Flat Hit@K} & \multicolumn{4}{|c}{Hierarchical precision@K}\\ \cline{3-11}
& K= & \text{1} & \text{2} & \text{5} & \text{10} & \text{20} & \text{2} & \text{5} & \text{10} & \text{20} \\ \hline 
\emph{2-hop} & ConSE(1) & 9.0 & 12.9 & 20.8 & 28.3 & 38.1 & 21.1 & 22.2 & 24.8 & 28.1 \\ 
						 & ConSE(10) & 9.2 & 13.7 & 22.4 & 31.0 & 41.4 & 22.5 & 24.1 & 27.3 & 30.8 \\
						 & ConSE(10) \cite{NorouziMBSSFCD14} & 9.4 & 15.1 & 24.7 & 32.7 & 41.8 & 21.4 & 24.7 & 26.9 & 28.4 \\
						 & ConSE(1000) & 8.9 & 13.3 & 21.8 & 30.1 & 40.3 & 22.0 & 23.7 & 27.0 & 30.4 \\
						 & ConSE(CV) & 8.3 & 12.9 & 21.8 & 30.9 & 41.7 & 21.5 & 23.8 & 27.5 & 31.3 \\ \cline{2-11}
						 & Ours$^\textrm{o-vs-o}$ & \textbf{\color{red}10.5} & \textbf{\color{red}16.7} & \textbf{\color{red}28.6} & \textbf{\color{red}40.1} & \textbf{\color{red}52.0} & \textbf{\color{red}25.1} & \textbf{\color{red}27.7} & \textbf{\color{red}30.3} & \textbf{\color{red}32.1} \\
						 & Ours$^\textrm{struct}$ & 9.8 & 15.3 & 25.8 & 35.8 & 46.5 & 23.8 & 25.8 & 28.2 & 29.6 \\
\hline
\emph{3-hop} & ConSE(1) & 2.8 & 4.2 & 7.2 & 10.1 & 14.3 & 6.2 & 18.4 & 20.4 & 22.1 \\ 
						 & ConSE(10) & \textbf{\color{red}2.9} & 4.5 & 7.7 & 11.3 & 16.1 & 6.9 & 20.9 & 23.1 & 25.2 \\
						 & ConSE(10) \cite{NorouziMBSSFCD14} & 2.7 & 4.4 & 7.8 & 11.5 & 16.1 & 5.3 & 20.2 & 22.4 & 24.7 \\
						 & ConSE(1000) & 2.8 & 4.3 & 7.4 & 10.9 & 15.6 & 6.8 & 20.7 & 22.9 & 25.1 \\
						 & ConSE(CV) & 2.6 & 4.1 & 7.3 & 11.1 & 16.4 & 6.7 & 21.4 & 23.8 & 26.3 \\ \cline{2-11}
						 & Ours$^\textrm{o-vs-o}$ & \textbf{\color{red}2.9} & \textbf{\color{red}4.9} & \textbf{\color{red}9.2} & \textbf{\color{red}14.2} & \textbf{\color{red}20.9} & 7.4 & \textbf{\color{red}23.7} & \textbf{\color{red}26.4} & \textbf{\color{red}28.6} \\
						 & Ours$^\textrm{struct}$ & \textbf{\color{red}2.9} & 4.7 & 8.7 & 13.0 & 18.6 & \textbf{\color{red}8.0} & 22.8 & 25.0 & 26.7 \\
\hline
\emph{All} & ConSE(1) & 1.4 & 2.3 & 3.8 & 5.6 & 7.8 & 3.0 & 7.6 & 8.7 & 9.6 \\ 
						 & ConSE(10) & \textbf{\color{red}1.5} & 2.3 & 4.0 & 6.0 & 8.7 & 3.3 & 8.9 & 10.2 & 11.4 \\
						 & ConSE(10) \cite{NorouziMBSSFCD14} & 1.4 & 2.2 & 3.9 & 5.8 & 8.3 & 2.5 & 7.8 & 9.2 & 10.4 \\
						 & ConSE(1000) & \textbf{\color{red}1.5} & 2.3 & 3.9 & 5.8 & 8.4 & 3.2 & 8.8 & 10.2 & 11.3 \\
						 & ConSE(CV) & 1.3 & 2.1 & 3.8 & 5.8 & 8.7 & 3.2 & 9.2 & 10.7 & 12.0 \\ \cline{2-11}
						 & Ours$^\textrm{o-vs-o}$ & 1.4 & \textbf{\color{red}2.4} & \textbf{\color{red}4.5} & \textbf{\color{red}7.1} & \textbf{\color{red}10.9} & 3.1 & 9.0 & 10.9 & \textbf{\color{red}12.5} \\
						 & Ours$^\textrm{struct}$ & \textbf{\color{red}1.5} & \textbf{\color{red}2.4} & 4.4 & 6.7 & 10.0 & \textbf{\color{red}3.6} & \textbf{\color{red}9.6} & \textbf{\color{red}11.0} & 12.2 \\
\hline
\end{tabular}
\end{table*}

\subsection{Qualitative results}
\begin{figure*}
\centering
\includegraphics[width=0.85\textwidth]{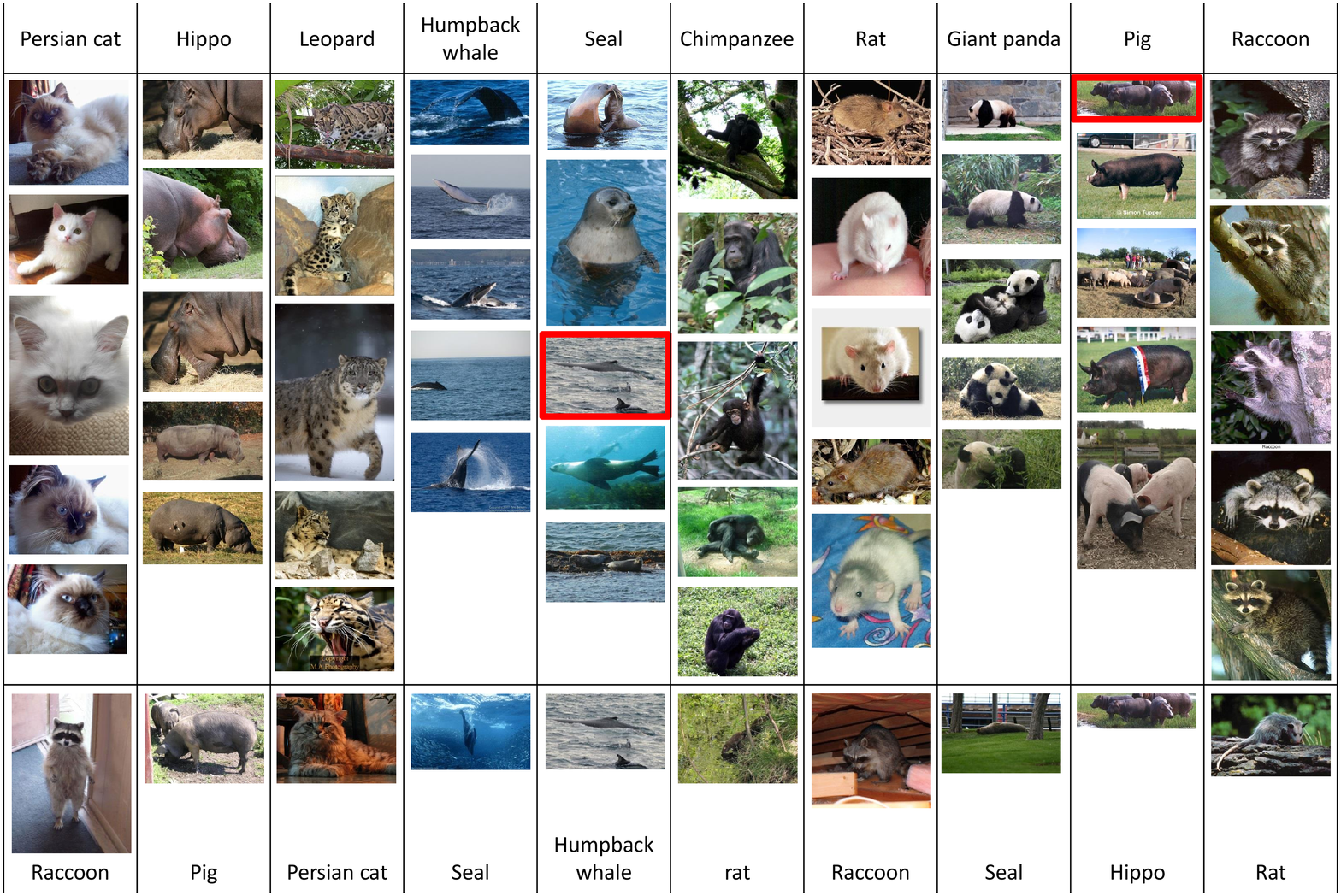}
\vskip -0.8em
\caption{\small Qualitative results of our method (Ours$^\textrm{struct}$) on \textbf{AwA}. \textbf{(Top)} We list the 10 unseen class labels. \textbf{(Middle)} We show the top-5 images classified into each class, according to the decision values. \emph{Misclassified images are marked with red boundaries}. \textbf{(Bottom)} We show the first misclassified image (according to the decision value) into each class and its ground-truth class label.}
\label{fig:AWA_1}
\end{figure*}

\begin{figure*}
\centering
\includegraphics[width=0.85\textwidth]{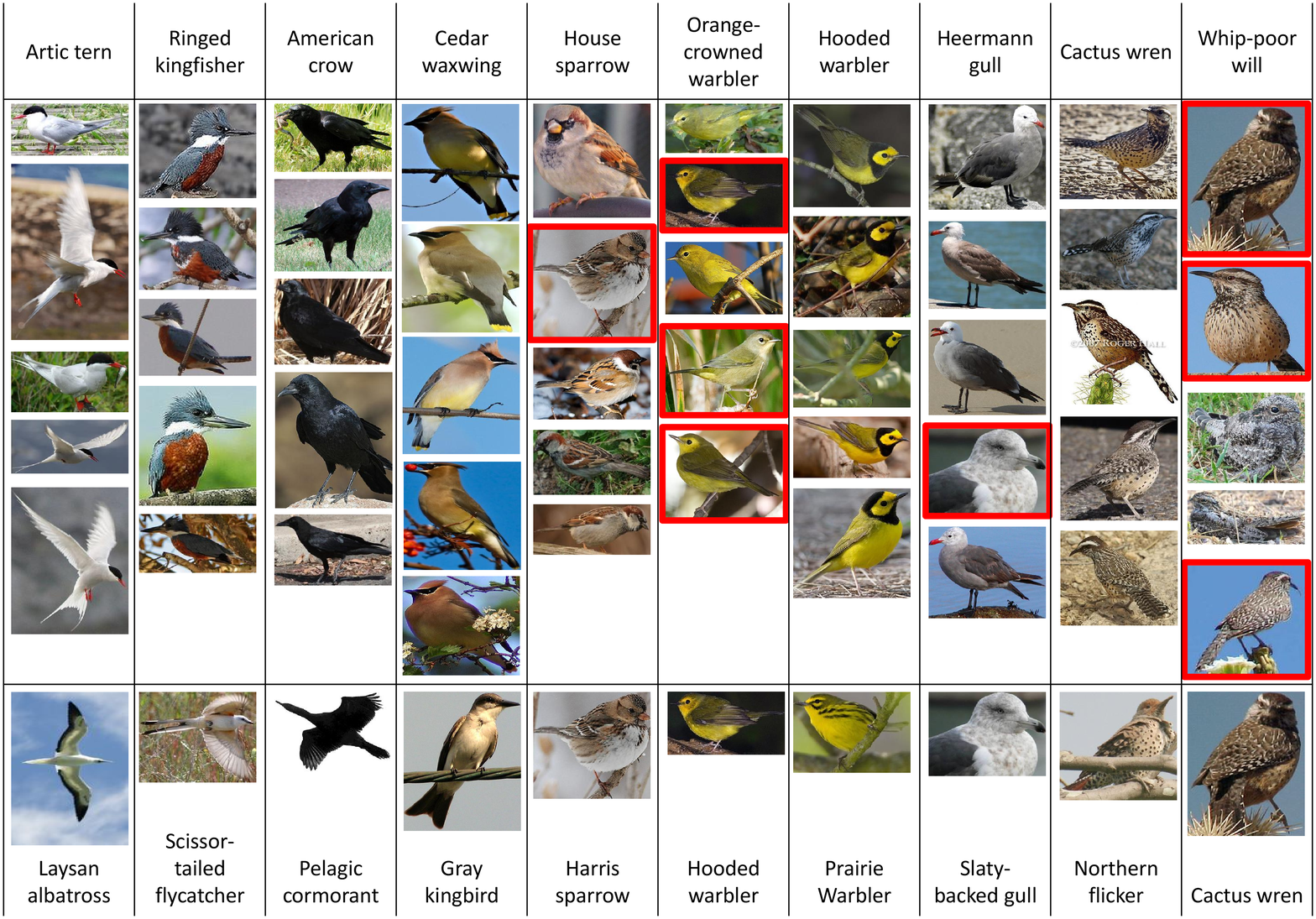}
\vskip -0.8em
\caption{\small Qualitative results of our method (Ours$^\textrm{struct}$) on \textbf{CUB}. \textbf{(Top)} We list a subset of unseen class labels. \textbf{(Middle)} We show the top-5 images classified into each class, according to the decision values. \emph{Misclassified images are marked with red boundaries}. \textbf{(Bottom)} We show the first misclassified image (according to the decision value) into each class and its ground-truth class label.}
\label{fig:CUB_1}
\end{figure*}

\begin{figure*}
\centering
\includegraphics[width=0.85\textwidth]{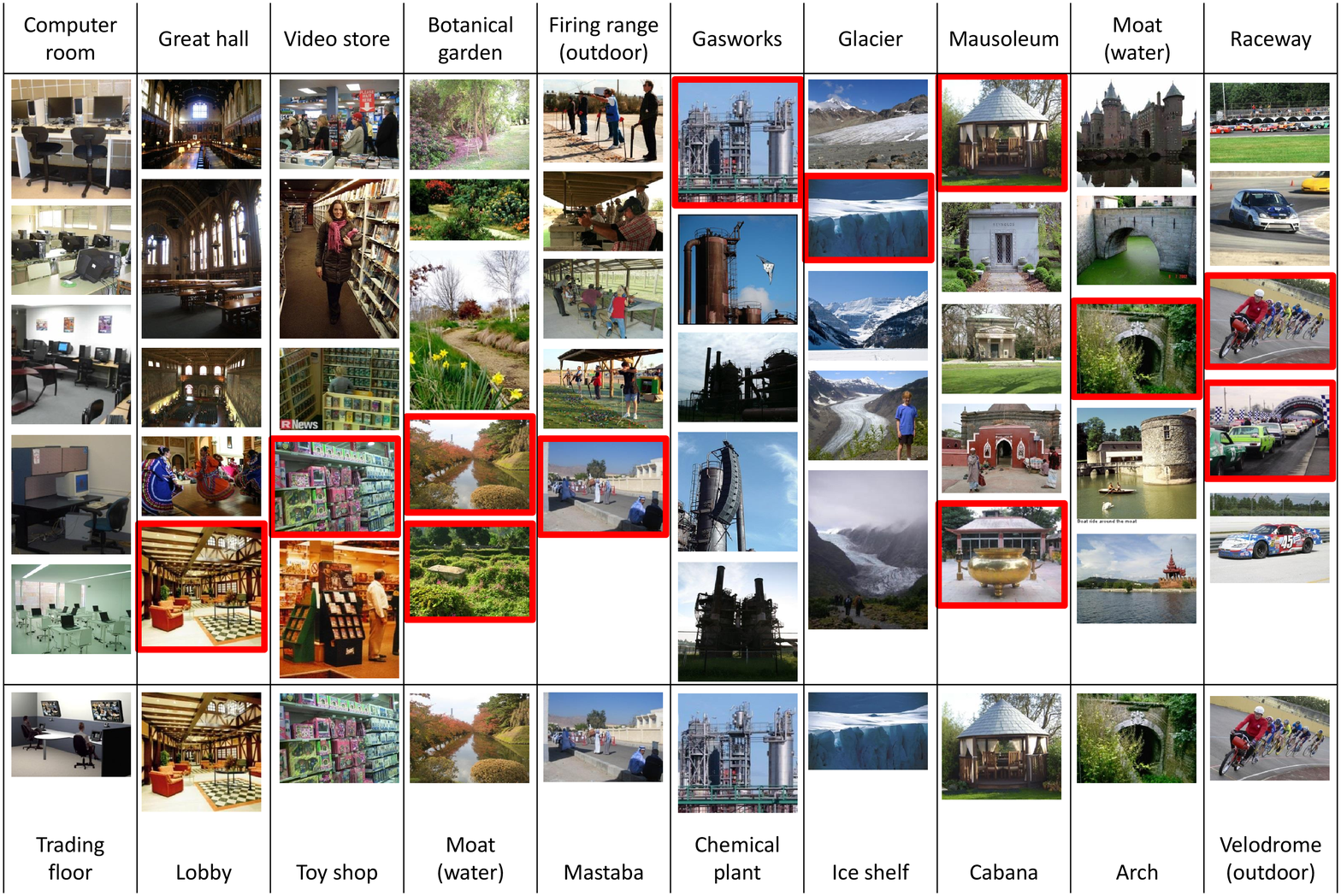}
\vskip -0.8em
\caption{\small Qualitative results of our method (Ours$^\textrm{o-vs-o}$) on \textbf{SUN}. \textbf{(Top)} We list a subset of unseen class labels. \textbf{(Middle)} We show the top-5 images classified into each class, according to the decision values. \emph{Misclassified images are marked with red boundaries}. \textbf{(Bottom)} We show the first misclassified image (according to the decision value) into each class and its ground-truth class label.}
\label{fig:SUN_1}
\end{figure*}

In this subsection, we present qualitative results of our method. We first illustrate what visual information the models (classifiers) for unseen classes capture, when provided with only semantic embeddings (no example images). In Figure~\ref{fig:AWA_1}, we list (on top) the 10 unseen class labels of \textbf{AwA}, and show (in the middle) the top-5 images classified into each class $c$, according to the decision values $\vw_c\T\vx$ (cf. eq. (1) and (4) of the main text). Misclassified images are marked with red boundaries. At the bottom, we show the first (highest score) misclassified image (according to the decision value) into each class and its ground-truth class label. According to the top images, our method reasonably captures discriminative visual properties of each unseen class based solely on its semantic embedding. We can also see that the misclassified images are with appearance so similar to that of predicted class that even humans cannot easily distinguish between the two. For example, the pig image at the bottom of the second column looks very similar to the image of hippos. Figure~\ref{fig:CUB_1} and Figure~\ref{fig:SUN_1} present the results in the same format on \textbf{CUB} and \textbf{SUN}, respectively (both on a subset of unseen class labels).

\begin{figure*}
\centering
\includegraphics[width=0.95\textwidth]{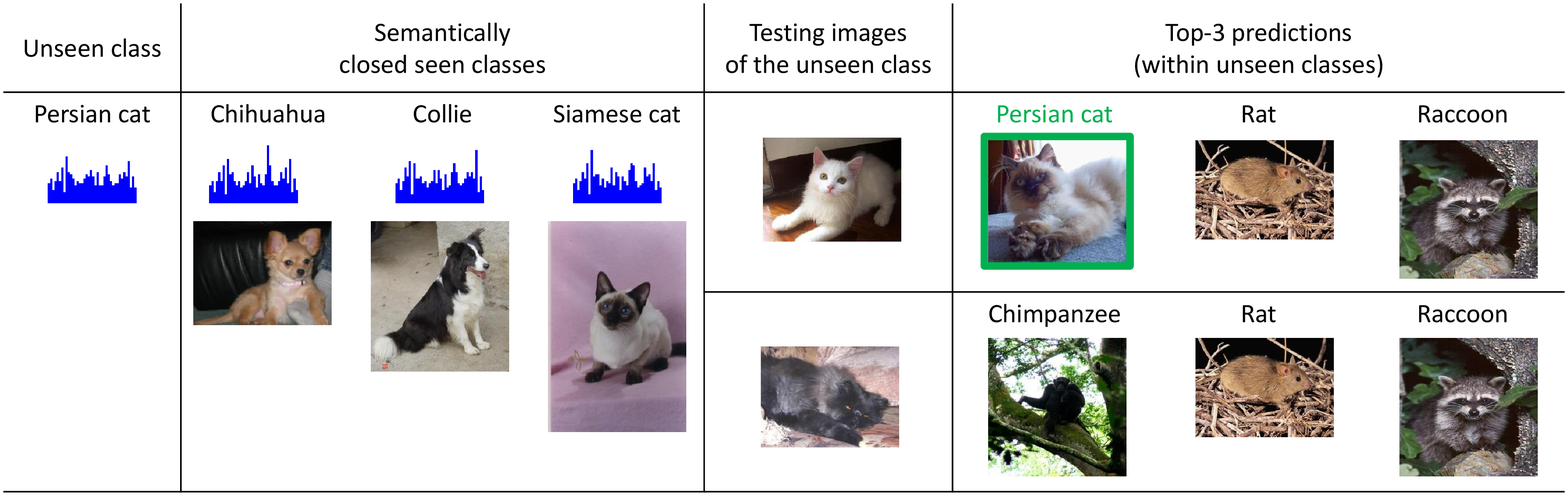}
\vskip -0.8em
\caption{\small Success/failure case analysis of our method (Ours$^\textrm{struct}$) on \textbf{AwA}: (Left) an unseen class label, (Middle-Left) the top-3 semantically similar seen classes to that unseen class, (Middle-Right) two test images of such unseen class, and (Right) the top-3 predicted unseen classes. The green text corresponds to the correct label.}
\label{fig:AWA_2}
\end{figure*}

\begin{figure*}
\centering
\includegraphics[width=0.95\textwidth]{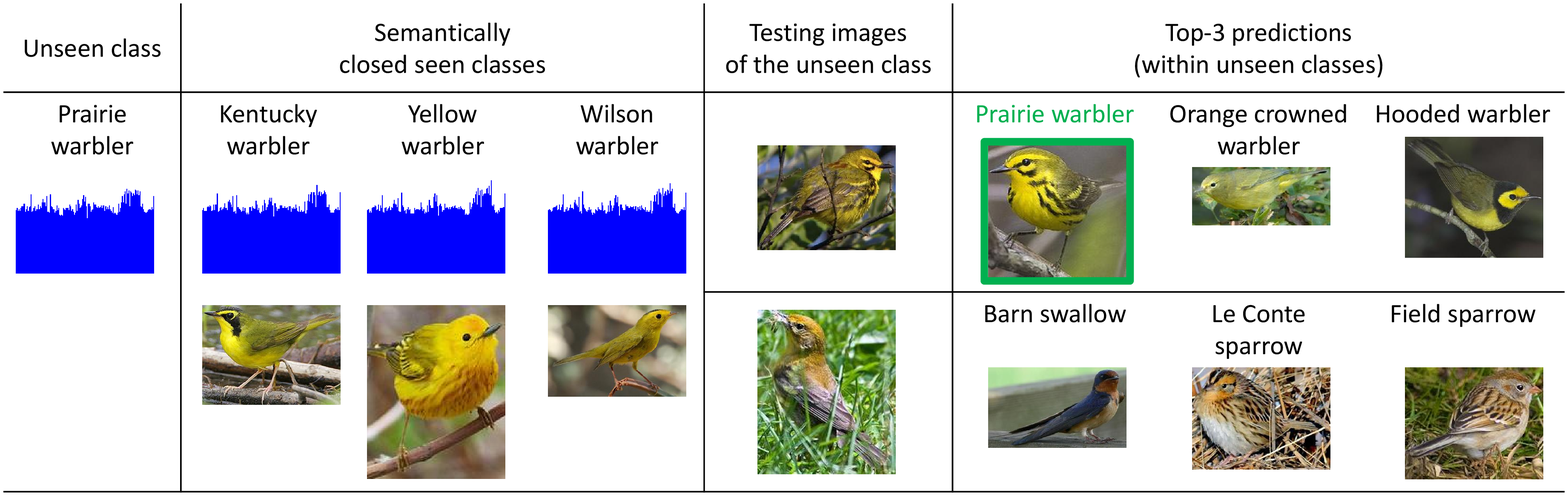}
\vskip -0.8em
\caption{\small Success/failure case analysis of our method (Ours$^\textrm{struct}$) on \textbf{CUB}. (Left) an unseen class label, (Middle-Left) the top-3 semantically similar seen classes to that unseen class, (Middle-Right) two test images of such unseen classes, and (Right) the top-3 predicted unseen class. The green text corresponds to the correct label.}
\label{fig:CUB_2}
\end{figure*}

\begin{figure*}
\centering
\includegraphics[width=0.95\textwidth]{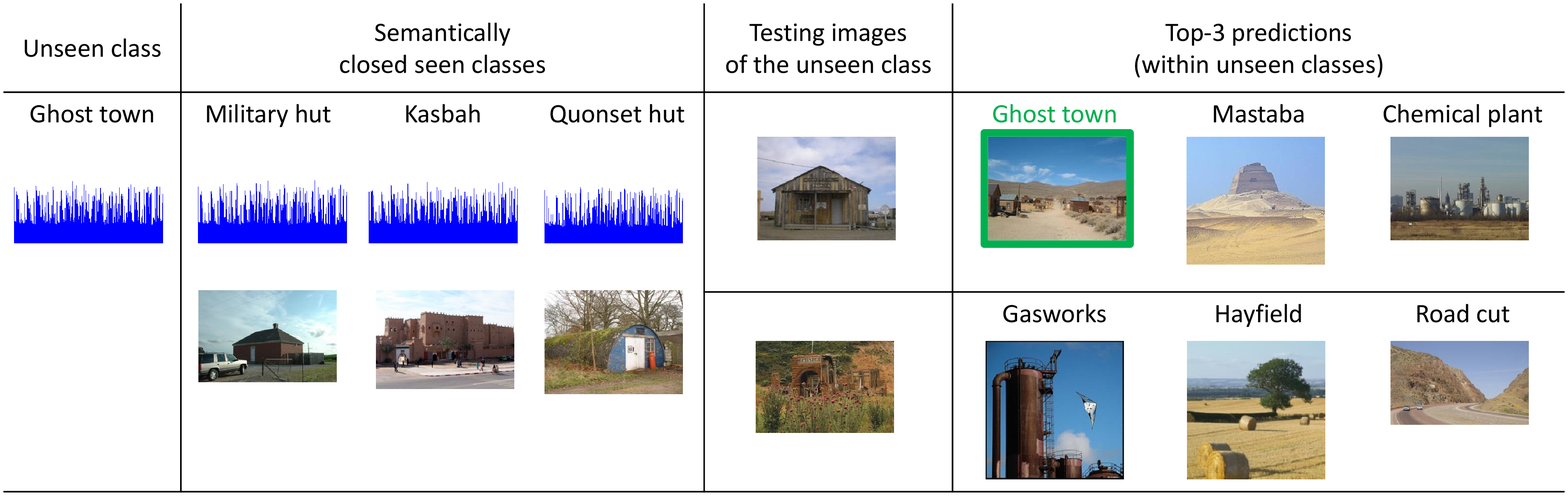}
\vskip -0.8em
\caption{\small Success/failure case analysis of our method (Ours$^\textrm{o-vs-o}$) on \textbf{SUN}. (Left) an unseen class label, (Middle-Left) the top-3 semantically similar seen classes to that unseen class, (Middle-Right) two test images of such unseen classes, and (Right) the top-3 predicted unseen class. The green text corresponds to the correct label.}
\label{fig:SUN_2}
\end{figure*}

We further analyze the success and failure cases; i.e., why an image from unseen classes is misclassified. The illustrations are in Figure~\ref{fig:AWA_2},~\ref{fig:CUB_2}, and~\ref{fig:SUN_2} for \textbf{AwA}, \textbf{CUB}, and \textbf{SUN}, respectively. In each figure, we consider \textbf{(Left)} one unseen class and show its convex combination weights $\vct{s}_c = \{s_{c1}, \cdots, s_{c\cR}\}$ as a histogram. We then present \textbf{(Middle-Left)} the top-3 semantically similar (in terms of $\vct{s}_c$) seen classes and their most representative images. As our model exploits phantom classes to connect seen and unseen classes in both semantic and model spaces, we expect that the model (classifier) for such unseen class captures similar visual information as those for semantically similar seen classes do. \textbf{(Middle-Right)} We examine two images of such unseen class, where the top one is correctly classified; the bottom one, misclassified. We also list \textbf{(Right)} the top-3 predicted labels (within the pool of unseen classes) and their most representative images. Green corresponds to correct labels. We see that, in the misclassified cases, the test images are visually dissimilar to those of the semantically similar seen classes. The synthesized unseen classifiers, therefore, cannot correctly recognize them, leading to incorrect predictions.

\subsection{Comparison between shallow and deep features of our approach}
In Table 4 of the main text, our approach performs better with deep features than with shallow features compared to other methods. We propose explanations for this phenomenal. Deep features are learned hierarchically and expected to be more abstract and semantically meaningful. Arguably, similarities between them (measured in inner products between classifiers) might be more congruent with similarities computed in the semantic embedding space for combining classifiers. Additionally, shallow features have higher dimensions (around 10,000) than deep features (e.g., 1024 for GoogLeNet) so they might require more phantom classes to synthesize classifiers. 

\subsection{Analysis on the number of base classifiers}

\begin{figure}[t]
\begin{center}
\includegraphics[width=0.9\columnwidth]{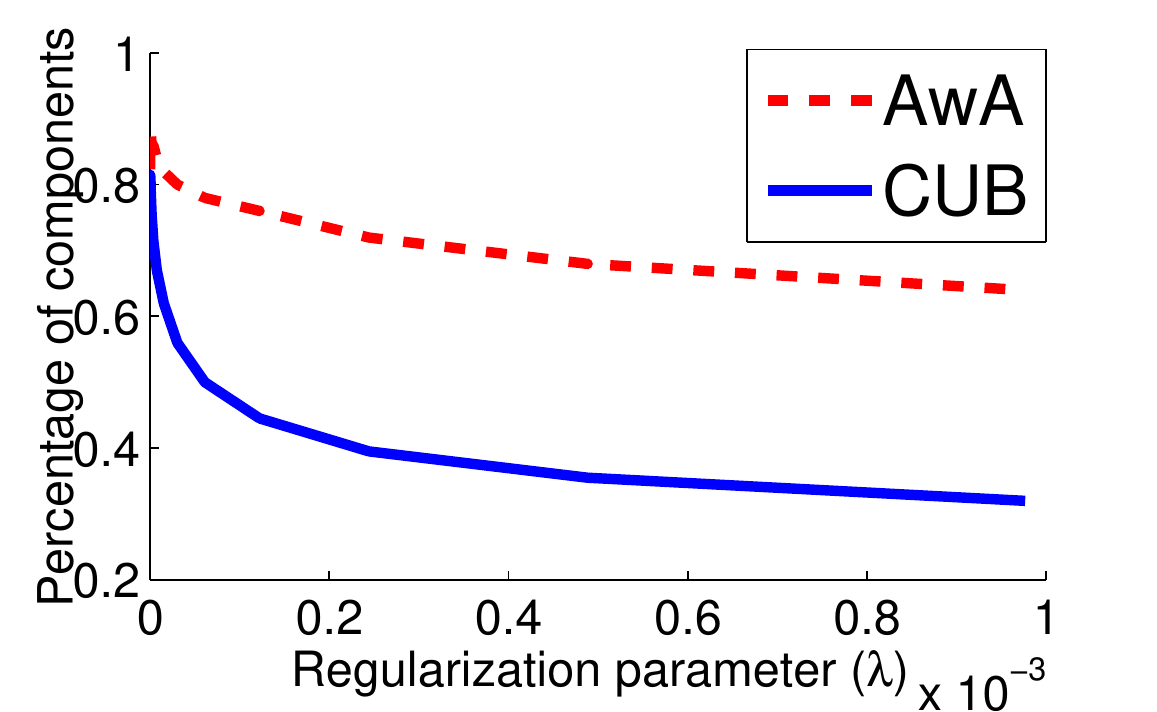}
\end{center}
\caption{\small Percentages of basis components required to capture 95\% of variance in classifier matrices for \textbf{AwA} and \textbf{CUB}.}
\label{fBasis}
\vskip -1em  
\end{figure}

In Fig. 2 of the main text, we show that even by using fewer base (phantom) classifiers than the number of seen classes (e.g., around 60 \%), we get comparable or even better results, especially for \textbf{CUB}. We surmise that this is because \textbf{CUB} is a fine-grained recognition benchmark and has higher correlations among classes, and provide analysis in Fig.~\ref{fBasis} to justify this.

We train one-versus-other classifiers for each value of the regularization parameter (i.e., $\lambda$ in eq.~(5) of the main text) on both \textbf{AwA} and \textbf{CUB}, and then perform PCA on the resulting classifier matrices. We then plot the required number (in percentage) of PCA components to capture 95\% of variance in the classifiers. Clearly, \textbf{AwA} requires more. This explains why we see the drop in accuracy for \textbf{AwA} but not \textbf{CUB} in Fig. 2 of the main text when using even fewer base classifiers. Particularly, the low percentage for \textbf{CUB} in Fig.~\ref{fBasis} implies that fewer base classifiers are possible. 

\end{document}